\documentclass[10pt]{article}
\ifdefined\pdfinfoomitdate\pdfinfoomitdate=1\fi
\ifdefined\pdftrailerid\pdftrailerid{}\fi
\ifdefined\pdfsuppressptexinfo\pdfsuppressptexinfo=-1\fi

\usepackage{fix-cm}
\usepackage[T1]{fontenc}
\usepackage[utf8]{inputenc}
\DeclareUnicodeCharacter{2212}{\ensuremath{-}}
\DeclareUnicodeCharacter{00D7}{\ensuremath{\times}}
\DeclareUnicodeCharacter{2265}{\ensuremath{\geq}}
\DeclareUnicodeCharacter{2264}{\ensuremath{\leq}}
\DeclareUnicodeCharacter{00B0}{\textdegree}
\DeclareUnicodeCharacter{00B1}{\ensuremath{\pm}}
\DeclareUnicodeCharacter{0394}{\ensuremath{\Delta}}
\DeclareUnicodeCharacter{2019}{'}
\DeclareUnicodeCharacter{2713}{\ensuremath{\checkmark}}
\DeclareUnicodeCharacter{2717}{\ensuremath{\times}}

\usepackage{libertinus}
\usepackage[scaled=0.91]{sourcesanspro}
\usepackage[scaled=0.94]{zi4}
\usepackage[letterpaper,left=0.82in,right=0.82in,top=0.72in,bottom=0.78in,headheight=14pt,footskip=25pt]{geometry}
\usepackage{microtype}
\usepackage{setspace}
\AtBeginDocument{\fontsize{10.25}{13.3}\selectfont}

\usepackage[dvipsnames,table]{xcolor}
\definecolor{ReportInk}{HTML}{1F2933}
\definecolor{HeadingGraphite}{HTML}{243746}
\definecolor{MutedSlate}{HTML}{5B6670}
\definecolor{RuleGrey}{HTML}{CBD5DD}
\definecolor{SurfaceGrey}{HTML}{F5F7F8}
\definecolor{BenchmarkBlue}{HTML}{1F4E79}
\definecolor{ModelPurple}{HTML}{654B9B}
\definecolor{RuntimeTeal}{HTML}{0B6E75}
\definecolor{PassGreen}{HTML}{2E7D32}
\definecolor{WarningAmber}{HTML}{8A5A00}
\definecolor{FailureRed}{HTML}{A33A3A}
\color{ReportInk}

\usepackage{titlesec}
\titleformat{\section}
  {\fontsize{15.5}{19}\selectfont\sffamily\bfseries\color{HeadingGraphite}}
  {\thesection}{0.62em}{}
\titleformat{\subsection}
  {\fontsize{12.5}{15.5}\selectfont\sffamily\bfseries\color{HeadingGraphite}}
  {\thesubsection}{0.58em}{}
\titleformat{\subsubsection}
  {\fontsize{10.5}{13.2}\selectfont\sffamily\bfseries\color{HeadingGraphite}}
  {\thesubsubsection}{0.52em}{}
\titlespacing*{\section}{0pt}{28pt plus 3pt minus 2pt}{8pt}
\titlespacing*{\subsection}{0pt}{20pt plus 2pt minus 1pt}{6pt}
\titlespacing*{\subsubsection}{0pt}{12pt}{4pt}

\usepackage{fancyhdr}
\fancypagestyle{plain}{\fancyhf{}\fancyfoot[R]{\sffamily\fontsize{7.5}{9.3}\selectfont\color{MutedSlate}\thepage}}

\usepackage{amsmath,amssymb}
\usepackage{booktabs,longtable,array,tabularx,multirow,colortbl}
\usepackage{ragged2e}
\usepackage{float}
\usepackage{siunitx}
\usepackage{calc}

\newcolumntype{Y}{>{\centering\arraybackslash}X}
\newcolumntype{R}{>{\raggedleft\arraybackslash}X}
\renewcommand{\arraystretch}{1.22}
\arrayrulecolor{RuleGrey}
\usepackage{enumitem}
\setlist{topsep=4pt,itemsep=1.5pt,parsep=0pt,leftmargin=1.45em}

\usepackage{graphicx}
\graphicspath{{figures/}}
\usepackage{caption}
\DeclareCaptionFont{captiontext}{\fontsize{8.5}{10.8}\selectfont}
\usepackage[section]{placeins}

\usepackage[most]{tcolorbox}
\newtcolorbox{abstractpanel}{
  enhanced,breakable,sharp corners,
  colback=SurfaceGrey,colframe=SurfaceGrey,boxrule=0pt,
  borderline west={2.2pt}{0pt}{HeadingGraphite},
  left=12pt,right=12pt,top=10pt,bottom=10pt,
  before skip=0pt,after skip=12pt
}
\newtcolorbox{statcardbox}[1]{
  enhanced,sharp corners,arc=1.5pt,
  colback=white,colframe=RuleGrey,boxrule=0.6pt,
  borderline north={2.5pt}{0pt}{#1},
  left=11pt,right=8pt,top=8pt,bottom=7pt,
  height=50pt,valign=center
}
\newcommand{\StatCard}[4]{%
  \begin{statcardbox}{#1}%
  {\sffamily\bfseries\fontsize{17.5}{18.5}\selectfont\color{#1}#2}\par
  {\sffamily\bfseries\fontsize{8.3}{10}\selectfont #3}%
  \if\relax\detokenize{#4}\relax\else\par{\sffamily\fontsize{7.5}{9.2}\selectfont\color{MutedSlate}#4}\fi
  \end{statcardbox}%
}
\newcommand{\FigureNote}[1]{%
  \par\vspace{4pt}\begingroup\RaggedRight\noindent{\sffamily\fontsize{7.5}{9.3}\selectfont\color{MutedSlate}\textbf{Note.} #1}\par\endgroup
}
\newcommand{\Code}[1]{{\ttfamily\fontsize{8.3}{10.3}\selectfont #1}}
\newcommand{\Artifact}[1]{{\ttfamily\fontsize{8.3}{10.3}\selectfont\Urlmuskip=0mu plus 1mu\relax #1}}
\usepackage{seqsplit}

\newcommand{\StatusPass}{\tcbox[on line,box align=base,sharp corners,arc=1pt,colback=PassGreen!8,colframe=PassGreen,boxrule=0.5pt,left=3pt,right=3pt,top=1pt,bottom=1pt]{\sffamily\bfseries\fontsize{7.2}{8.6}\selectfont\color{PassGreen}\ensuremath{\checkmark} PASS}}
\newcommand{\StatusFail}{\tcbox[on line,box align=base,sharp corners,arc=1pt,colback=FailureRed!7,colframe=FailureRed,boxrule=0.5pt,left=3pt,right=3pt,top=1pt,bottom=1pt]{\sffamily\bfseries\fontsize{7.2}{8.6}\selectfont\color{FailureRed}\ensuremath{\times} FAIL}}
\newcommand{\ObjectNote}[1]{%
  \par\vspace{4pt}\begingroup\RaggedRight\noindent{\sffamily\fontsize{7.5}{9.3}\selectfont\color{MutedSlate}\textbf{Note.} #1}\par\endgroup
}
\newtcolorbox{reportingnotebox}{
  enhanced,breakable,sharp corners,
  colback=SurfaceGrey,colframe=SurfaceGrey,boxrule=0pt,
  borderline west={2.2pt}{0pt}{HeadingGraphite},
  left=10pt,right=10pt,top=8pt,bottom=8pt,
  before skip=12pt,after skip=12pt,
  fontupper=\sffamily\fontsize{8.5}{10.8}\selectfont
}
\newtcolorbox{evidenceboundarybox}{
  enhanced,breakable,sharp corners,
  colback=WarningAmber!6,colframe=WarningAmber!24,boxrule=0.6pt,
  borderline west={2.4pt}{0pt}{WarningAmber},
  title={Evidence boundary},
  coltitle=WarningAmber,fonttitle=\sffamily\bfseries\fontsize{8.5}{10.5}\selectfont,
  left=10pt,right=10pt,top=7pt,bottom=8pt,
  before skip=12pt,after skip=12pt,
  fontupper=\fontsize{9.2}{12}\selectfont
}
\newcommand{\TableText}{\sffamily\fontsize{8.5}{10.5}\selectfont}

\newtcolorbox{policybox}{
  enhanced,breakable,sharp corners,
  colback=SurfaceGrey,colframe=RuleGrey,boxrule=0.55pt,
  borderline west={2.4pt}{0pt}{HeadingGraphite},
  left=12pt,right=12pt,top=10pt,bottom=10pt,
  before skip=12pt,after skip=14pt
}
\newcommand{\PolicyKicker}[1]{%
  \noindent{\sffamily\bfseries\fontsize{8}{9.8}\selectfont\color{HeadingGraphite}\MakeUppercase{#1}}\par\vspace{6pt}%
}
\newtcolorbox{manifestbox}{
  enhanced,breakable,sharp corners,
  colback=SurfaceGrey,colframe=RuleGrey,boxrule=0.55pt,
  borderline west={2.4pt}{0pt}{BenchmarkBlue},
  left=12pt,right=12pt,top=9pt,bottom=9pt,
  before skip=10pt,after skip=12pt
}
\newcommand{\HashTwo}[2]{%
  \begin{tabular}[t]{@{}l@{}}\ttfamily\fontsize{7.75}{9.4}\selectfont\hyphenchar\font=-1 #1\\#2\end{tabular}%
}

\usepackage{xurl}
\usepackage[numbers,sort&compress]{natbib}
\usepackage{hyperref}
\hypersetup{
  colorlinks=true,
  linkcolor=HeadingGraphite,
  citecolor=ModelPurple,
  urlcolor=RuntimeTeal,
  pdftitle={CropCop: An Auditable 120-Class Plant-Health Model from Benchmark Reconstruction to a Quantised Runtime Artifact},
  pdfauthor={Rana Muhammad Ahmed; Sabahat Abbas},
  pdfsubject={Plant-health recognition, benchmark forensics, quantisation, and runtime verification},
  pdfkeywords={plant-health recognition, dataset leakage, duplicate forensics, DINOv3, MobileNetV4, quantisation, ExecuTorch, XNNPACK}
}
\newcommand{\papertitle}{CropCop: An Auditable 120-Class Plant-Health Model from Benchmark Reconstruction to a Quantised Runtime Artifact}
\newcommand{\paperauthors}{Rana Muhammad Ahmed\textsuperscript{1} \quad Sabahat Abbas\textsuperscript{1}}

\newcommand{\paperaffiliation}{\textsuperscript{1}Department of Computer Science, Bahria University Islamabad, Islamabad, Pakistan}
\newcommand{\paperdate}{August 2026}

\begin{document}
\thispagestyle{empty}

\noindent{\sffamily\fontsize{8}{10}\selectfont\bfseries\color{BenchmarkBlue}\MakeUppercase{Research Article \enspace\textperiodcentered\enspace Computer Vision for Plant Health}}\par
\vspace{8pt}
\noindent{\color{RuleGrey}\rule{\linewidth}{0.7pt}}\par
\vspace{12pt}
{\raggedright\fontsize{24}{27.5}\selectfont\sffamily\bfseries\color{HeadingGraphite}\papertitle\par}
\vspace{10pt}
{\raggedright\fontsize{11.5}{14}\selectfont\sffamily\bfseries\paperauthors\par}
\vspace{3pt}
{\raggedright\fontsize{8.8}{11}\selectfont\sffamily\color{MutedSlate}\paperaffiliation\par}
\vspace{3pt}
{\raggedright\fontsize{8.8}{11}\selectfont\sffamily\color{MutedSlate}Preprint \textperiodcentered\ \paperdate\par}
\vspace{14pt}

\begin{abstractpanel}
{\sffamily\bfseries\fontsize{10.5}{13.2}\selectfont\color{HeadingGraphite}Abstract}\par\vspace{4pt}
{\fontsize{9.5}{12.6}\selectfont
A plant-health score can appear precise while resting on duplicated image families, a long-tailed label space, or a runtime file that was never evaluated. We present CropCop, a closed-set recognition system spanning 120 operational plant-health classes and an evidence chain from corpus reconstruction to direct execution of the final quantised artifact. Starting from 117,546 audited images, we rejected the inherited partition after confirming 3,233 duplicate relationships across split boundaries and froze a 109,107-image benchmark with zero crossings among the audited trusted leakage groups and a 151.7$\times$ largest-to-smallest class ratio. A fully fine-tuned DINOv3 ConvNeXt-Tiny reference achieved 98.51\% accuracy and 96.87\% macro-F1 on the locked internal test. A compact MobileNetV4 Conv-Medium derivative achieved 98.46\% accuracy and 96.27\% macro-F1 without being presented as evidence for a new distillation method. Validation-only post-training quantisation selected dynamic activations with per-channel weights, and the final 22.60 MiB ExecuTorch/XNNPACK PTE achieved 98.46\% accuracy and 96.23\% macro-F1 when executed directly. Only six of 16,363 top-1 decisions changed between the converted INT8 graph and the PTE, while paired analysis showed a modest class-balanced loss; an exploratory post hoc fruit-label slice localized a larger recall decline than aggregate accuracy revealed. CropCop establishes strong leakage-controlled internal recognition and software-runtime fidelity; it does not establish performance on unseen farms, camera pipelines, or physical Android hardware.}
\end{abstractpanel}

\noindent
\begin{tabularx}{\linewidth}{@{}X@{\hspace{8pt}}X@{\hspace{8pt}}X@{\hspace{8pt}}X@{}}
\StatCard{BenchmarkBlue}{109,107}{frozen images}{} &
\StatCard{BenchmarkBlue}{120}{operational classes}{} &
\StatCard{RuntimeTeal}{22.60 MiB}{executed PTE}{} &
\StatCard{RuntimeTeal}{6 / 16,363}{changed top-1 decisions}{} \\
\end{tabularx}
\vspace{8pt}

\noindent{\sffamily\fontsize{8.5}{10.5}\selectfont\bfseries\color{HeadingGraphite}Keywords}\quad
{\fontsize{8.8}{11}\selectfont plant-health recognition; dataset leakage; duplicate forensics; long-tailed classification; DINOv3; MobileNetV4; post-training quantisation; ExecuTorch; XNNPACK; runtime fidelity}
\vspace{10pt}

\section{Introduction}\label{introduction}

The impressive number in a plant-disease paper is usually the accuracy. The consequential numbers are often elsewhere: how many test images have relatives in training, how many classes have only a handful of examples, and how many predictions change after export. Plant-health recognition is attractive because cameras are inexpensive and already embedded in agricultural workflows, but a defensible result requires more than fitting a classifier. The benchmark must have a stable identity, the evaluation must resist selection leakage, and the file intended for inference must be shown to behave like the model described in the paper.

Plant-image collections make these requirements difficult. Public corpora are assembled from controlled repositories, internet searches, extension websites, derivative datasets, and field photographs. The resulting images vary in background, framing, compression, resolution, plant organ, cultivar, disease severity, and diagnostic specificity. Byte-identical copies are only the simplest source of contamination. Crops, recompressions, colour transformations, and burst-like views can cross nominal train and test boundaries, while label-correlated borders, dimensions, and acquisition styles can remain predictive after duplicate removal. PlantVillage demonstrated that deep networks could classify controlled imagery with striking internal accuracy, yet its external-image experiment also revealed how sharply performance can change across acquisition domains \citep{hughes2015plantvillage, mohanty2016plant}. PlantDoc, PlantWild, PlantWild\_v2, PlantSeg, and Deep-Plant-Disease have since broadened the field toward natural imagery and larger taxonomies \citep{singh2020plantdoc, wei2024plantwild, wei2025plantwildv2, wei2026plantseg, chai2025deepplant}.

Compact inference creates a second identity problem. A fine-tuned checkpoint, a selected float student, a quantised graph, a backend-lowered program, and a serialised runtime file are different computational objects. A changed class map, preprocessing contract, unsupported operator, backend rewrite, or quantisation decision can alter predictions without changing the project name. Reporting the checkpoint score beside an unevaluated deployment claim conceals where the result was preserved or lost. CropCop therefore treats model conversion as a sequence of paired evaluations. Every state is tied to the same class order and locked rows, and the final runtime claim applies to one exact PTE identified by its bytes and SHA-256 digest.

The project began with 117,546 audited images and 121 candidate labels. The inherited folders were abandoned after 3,233 confirmed duplicate relationships crossed historical train, validation, and test boundaries. Global reconstruction produced CropCop Final v1: 109,107 images across 120 operational classes, divided into 76,376 training, 16,368 validation, and 16,363 test rows, with zero crossings among the audited trusted leakage groups. The dataset remains difficult. Its largest class is 151.7 times the size of its smallest; its labels mix specific diseases with broader condition categories; and its images range from isolated leaves and fruits to whole plants and field scenes.

\begin{figure}[H]
\centering
\includegraphics[width=\linewidth]{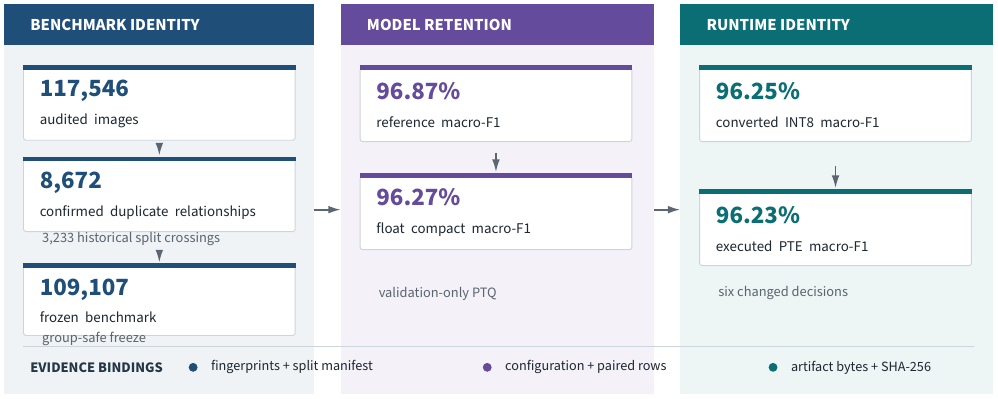}
\caption{CropCop audit-to-runtime evidence chain.}
\label{fig:evidence-chain}
\FigureNote{Each transition is tied to a persisted manifest, prediction record, configuration fingerprint, or artifact identifier.}
\end{figure}

CropCop asks three research questions. \textbf{RQ1 --- benchmark validity:} can a heterogeneous plant-image aggregate be reconstructed so that confirmed duplicate families do not cross the final partitions? \textbf{RQ2 --- model retention:} how much class-balanced performance remains when a strong DINOv3 reference is carried into a compact MobileNetV4 lineage? \textbf{RQ3 --- runtime fidelity:} does the selected quantised graph survive serialisation into an ExecuTorch/XNNPACK PTE without materially changing its decisions or ground-truth performance? The questions are deliberately narrower than ``does the model work in the field?'' That claim requires a source-independent cohort and is outside the completed evidence. Figure~\ref{fig:evidence-chain} summarises the connected evidence chain.

The completed system answers the three questions with a connected result. The DINOv3 ConvNeXt-Tiny reference reached 98.51\% accuracy and 96.87\% macro-F1 on the locked internal test. The MobileNetV4 float state retained 98.46\% accuracy and 96.27\% macro-F1. The final PTE reached 98.46\% accuracy and 96.23\% macro-F1; only six top-1 decisions differed from the converted INT8 graph. The contribution is not a new backbone, loss, or quantiser. It is an auditable model study that connects benchmark repair, class-balanced evaluation, compact transfer, validation-only PTQ selection, and direct runtime execution while preserving the boundary between internal evidence and external validity.

\section{Related Work}\label{related-work}

\subsection{Plant-health datasets and domain realism}\label{plant-health-datasets-and-domain-realism}

Early plant-recognition work benefited from controlled image collections whose clean backgrounds and centred leaves made large-scale supervised learning practical \citep{hughes2015plantvillage, mohanty2016plant}. Those datasets established the feasibility of image-based disease recognition, but they also made acquisition conditions part of the task. PlantDoc introduced field and internet imagery \citep{singh2020plantdoc}. PlantWild and its refined v2 release emphasised natural variation, fine-grained ambiguity, and broader label spaces \citep{wei2024plantwild, wei2025plantwildv2}. PlantSeg extended in-the-wild work to localisation and segmentation \citep{wei2026plantseg}, while Deep-Plant-Disease assembled a larger multi-crop identification benchmark \citep{chai2025deepplant}. These resources show why raw accuracy across different datasets is not a meaningful leaderboard: ontology, source composition, image granularity, duplicate policy, and split construction all change the question being measured.

\subsection{Benchmark contamination, shortcut learning, and reporting}\label{benchmark-contamination-shortcut-learning-and-reporting}

Near-duplicate contamination can alter benchmark estimates even in canonical vision datasets. The ciFAIR analysis showed that CIFAR train--test near-duplicates changed reported performance and motivated explicitly deduplicated variants \citep{barz2020cifair}. Label-error studies have likewise shown that test-set defects can destabilise comparisons \citep{northcutt2021labelerrors}, while representation-based audit methods treat duplicates, off-topic images, and suspect labels as distinct data-quality problems \citep{groger2024selfclean}. A leakage-free partition still does not guarantee causal learning. Models can exploit acquisition signatures that correlate with labels, as shortcut-learning studies in medical imaging have demonstrated \citep{degrave2021shortcuts}. Datasheets and model cards provide complementary reporting frameworks for documenting composition, intended use, limitations, and model identity \citep{gebru2021datasheets, mitchell2019modelcards}. CropCop applies these ideas to one agricultural corpus and links the audit directly to the downstream runtime lineage.

\subsection{Foundation transfer, long-tailed recognition, and compact students}\label{foundation-transfer-long-tailed-recognition-and-compact-students}

Self-supervised visual representations have raised the transfer baseline for tasks with limited or heterogeneous supervision. DINOv2 demonstrated broad transfer from self-supervised features \citep{oquab2024dinov2}, and DINOv3 extended that family to multiple architectures and resource points \citep{simeoni2025dinov3}. CropCop uses DINOv3 twice: a frozen ViT-S/16 representation diagnoses the dataset before task adaptation, and a ConvNeXt-Tiny model \citep{liu2022convnext} provides the fully fine-tuned reference. The dataset's 151.7× class-size ratio also places it in the long-tail regime, where class-balanced losses and logit adjustment have been proposed to reduce head-class dominance \citep{cui2019classbalanced, menon2021logitadjustment}. CropCop does not claim a new imbalance method. Macro-F1, balanced accuracy, disjoint support bins, and per-class tables are used to prevent aggregate accuracy from hiding minority behaviour.

Knowledge distillation transfers information from a teacher into a smaller model \citep{hinton2015distilling, gou2021kdsurvey}, and recent plant-recognition work has studied distillation as an explicit efficiency tool \citep{moummad2026efficient}. MobileNetV4 was designed for hardware-aware performance across diverse mobile backends \citep{qin2024mobilenetv4}, while recent plant benchmarks have compared lightweight architectures across broad taxonomies \citep{kumar2025mobileplant}. Quantisation methods reduce numerical precision to lower storage and execution costs \citep{jacob2018quantization}. Calibration distinguishes predictive correctness from confidence quality \citep{guo2017calibration}. ExecuTorch then introduces further states---export, backend lowering, quantisation, serialisation, and runtime execution---with XNNPACK-specific quantisation support \citep{executorch2026export, executorch2026xnnpack}. CropCop's position across these lines of work is summarised in Table~\ref{tab:literature-positioning}.

\begingroup\TableText\setlength{\tabcolsep}{3.2pt}
\begin{longtable}{@{}>{\RaggedRight\arraybackslash}p{0.235\linewidth}>{\RaggedRight\arraybackslash}p{0.205\linewidth}>{\RaggedRight\arraybackslash}p{0.235\linewidth}>{\RaggedRight\arraybackslash}p{0.265\linewidth}@{}}
\caption{Positioning of CropCop relative to adjacent literature.}\label{tab:literature-positioning}\\
\rowcolor{HeadingGraphite}
\color{white}\sffamily\bfseries Literature line &
\color{white}\sffamily\bfseries Primary object &
\color{white}\sffamily\bfseries What it establishes &
\color{white}\sffamily\bfseries Gap relative to CropCop's question \\
\endfirsthead
\multicolumn{4}{r}{\sffamily\fontsize{7.5}{9.3}\selectfont\color{MutedSlate}Table~\thetable\ --- continued}\\[4pt]
\rowcolor{HeadingGraphite}
\color{white}\sffamily\bfseries Literature line &
\color{white}\sffamily\bfseries Primary object &
\color{white}\sffamily\bfseries What it establishes &
\color{white}\sffamily\bfseries Gap relative to CropCop's question \\
\endhead
\bottomrule
\endfoot
\bottomrule
\endlastfoot
\rowcolor{SurfaceGrey}
Controlled and in-the-wild plant benchmarks \citep{hughes2015plantvillage, mohanty2016plant, singh2020plantdoc, wei2024plantwild, wei2025plantwildv2, wei2026plantseg, chai2025deepplant} & Dataset and recognition task & Scale, taxonomy, acquisition difficulty, or localisation & The benchmark is normally the endpoint; compact runtime identity is not the central object. \\
Benchmark audit and reporting \citep{barz2020cifair, northcutt2021labelerrors, groger2024selfclean, degrave2021shortcuts, gebru2021datasheets, mitchell2019modelcards} & Contamination, labels, shortcuts, documentation & Why data identity, provenance, and limitations matter & These works are domain-general and do not trace one agricultural model through quantised execution. \\
\rowcolor{SurfaceGrey}
Foundation transfer and long-tail learning \citep{oquab2024dinov2, simeoni2025dinov3, liu2022convnext, cui2019classbalanced, menon2021logitadjustment} & Representations and class imbalance & Transfer baselines and class-balanced objectives & They do not by themselves establish benchmark integrity or runtime fidelity. \\
Distillation and efficient vision \citep{hinton2015distilling, gou2021kdsurvey, moummad2026efficient, qin2024mobilenetv4, kumar2025mobileplant, jacob2018quantization} & Compact-model training and quantisation & How accuracy can be transferred or compressed & Dataset forensics and exact serialized-artifact evaluation are usually separate concerns. \\
\rowcolor{SurfaceGrey}
Calibration and runtime tooling \citep{guo2017calibration, executorch2026export, executorch2026xnnpack} & Confidence and executable backends & How predictions can be calibrated, exported, lowered, and quantised & They supply methods and infrastructure rather than an end-to-end application evidence chain. \\
\rowcolor{BenchmarkBlue!8}
\textbf{CropCop} & One audited plant-health model lineage & Benchmark reconstruction, locked evaluation, compact transfer, PTQ selection, direct PTE execution, and artifact identity & The contribution is the connected evidence chain, not a new backbone and not a claim beyond the evaluated distribution. \\
\end{longtable}
\endgroup

The table is an evidence-scope comparison, not a claim that earlier work should have implemented every stage. CropCop's contribution lies in making the transitions jointly observable for one model. The cited literature supplies the data, learning, calibration, and runtime foundations; the present study tests whether their outputs can remain attached to one identifiable result.

\section{Dataset and Benchmark Construction}\label{dataset-and-benchmark-construction}

\subsection{Source pool and operational ontology}\label{source-pool-and-operational-ontology}

The audited source pool contained 117,546 images arranged under 121 candidate labels. The labels were operational categories rather than a uniform botanical taxonomy: some named specific diseases, others described healthy states, broad fruit conditions, nutrient deficiency, or a generic \Code{plant\_healthy} category. Images included leaves, fruits, whole plants, close-ups, and field scenes. This heterogeneity is part of the completed task and one reason the paper uses the phrase \emph{plant-health recognition} rather than claiming 120 distinct diseases.

The frozen class map contains 120 operational labels spanning 35 named crop or plant groups; \Code{nutrient\_deficiency} and \Code{plant\_healthy} do not encode a crop. Under the literal label structure, 72 labels name a specific disease or named condition, 9 use a broad condition category, and 39 denote a healthy state. The broad group comprises generic fruit- or leaf-disease labels, \Code{bean\_fungal\_disease}, \Code{corn\_fungal\_leaf}, and \Code{nutrient\_deficiency}. This breakdown substantiates the ontology boundary: the task is a closed-set plant-health vocabulary rather than a uniform set of 120 disease entities. The full class inventory and split supports are reported in Appendix~\ref{operational-class-taxonomy-and-support-inventory}.

\begin{table}[H]
\centering
\caption{Operational class-map composition.}
\label{tab:taxonomy-summary}
\begingroup
\TableText
\setlength{\tabcolsep}{7pt}
\renewcommand{\arraystretch}{1.28}
\begin{tabularx}{0.82\linewidth}{@{}>{\RaggedRight\arraybackslash}p{0.27\linewidth}>{\RaggedRight\arraybackslash}Xr@{}}
\rowcolor{BenchmarkBlue}
\color{white}\bfseries Dimension & \color{white}\bfseries Category & \color{white}\bfseries Count \\
\rowcolor{BenchmarkBlue!8}
Crop scope & Named crop or plant groups & 35 \\
\rowcolor{SurfaceGrey}
Crop scope & Crop-unspecified labels & 2 \\
Label type & Specific named disease\textsuperscript{\textdagger} & 72 \\
\rowcolor{SurfaceGrey}
Label type & Broad condition category & 9 \\
Label type & Generic healthy state & 39 \\
\rowcolor{BenchmarkBlue!8}
Label type & \textbf{Total operational classes} & \textbf{120} \\
\end{tabularx}
\endgroup
\ObjectNote{Crop/species and label type are derived from the literal operational labels. The four pest-labelled classes marked in Appendix E are retained in the requested specific-named-disease category because the three-way scheme has no separate pest category.}
\end{table}

The historical assignment placed 57,426 images in training, 12,272 in validation, and 47,848 in test. Those folders were retained as provenance, not accepted as a scientific split. The audit treated image identity as a graph problem. Nodes represented files; trusted edges represented direct evidence that two files belonged to the same image family. Edge sources included exact identity, strong hash evidence, and corrected feature/geometric verification. Cross-label edges were handled separately because a duplicate family carrying different labels creates an annotation conflict rather than an ordinary redundancy.

\subsection{Why the inherited partition was rejected}\label{why-the-inherited-partition-was-rejected}

The audit found 8,672 trusted duplicate relationships. Of these, 3,233 crossed historical split boundaries: 2,876 exact and 357 near-duplicate relationships. This was not a marginal defect. A model evaluated on the inherited folders could encounter a transformed or byte-identical member of a training image family in its test set. The historical split was therefore discarded before the final model lineage was selected. The intervention changes the semantics of the test: it no longer rewards known duplicate-family overlap, although it still represents the reconstructed aggregate rather than an independent acquisition domain.

\subsection{Corrected near-duplicate verification}\label{corrected-near-duplicate-verification}

An earlier feature route used an ORB-derived score whose denominator could produce invalid ratios above one. Rather than preserving its decisions, the reconstruction reopened all 2,031 candidate edges that depended on that route. The corrected ORB/RANSAC verifier limited the maximum image side to 800 pixels, extracted up to 1,200 ORB features, required Hamming-compatible matches and a normalised good-match ratio of at least 0.12, and accepted a geometric relation only when homography support, inlier ratio, spatial coverage, and reprojection error passed fixed thresholds. The corrected route retained 1,932 edges and rejected 99. This history matters because an audit is not made reliable by hiding a defective detector; reliability comes from reopening the affected decisions and recording the correction.

\subsection{Trusted graph, direct-cover removal, and human review}\label{trusted-graph-direct-cover-removal-and-human-review}

The final trusted graph contained 6,196 exact edges, 445 strong-hash edges, 1,932 corrected feature edges, and 17 confirmed cross-label edges. Deduplication used a direct-cover rule: every removed image required a direct trusted relation to a retained representative. This is more conservative than deleting an entire transitive component when some links may be uncertain. The rule removed 8,355 duplicate files. Thirty-four images in unresolved cross-label families, six severe-quality failures, and the one-image \Code{rice\_neck\_blast} category were quarantined.

\begin{figure}[H]
\centering
\includegraphics[width=\linewidth]{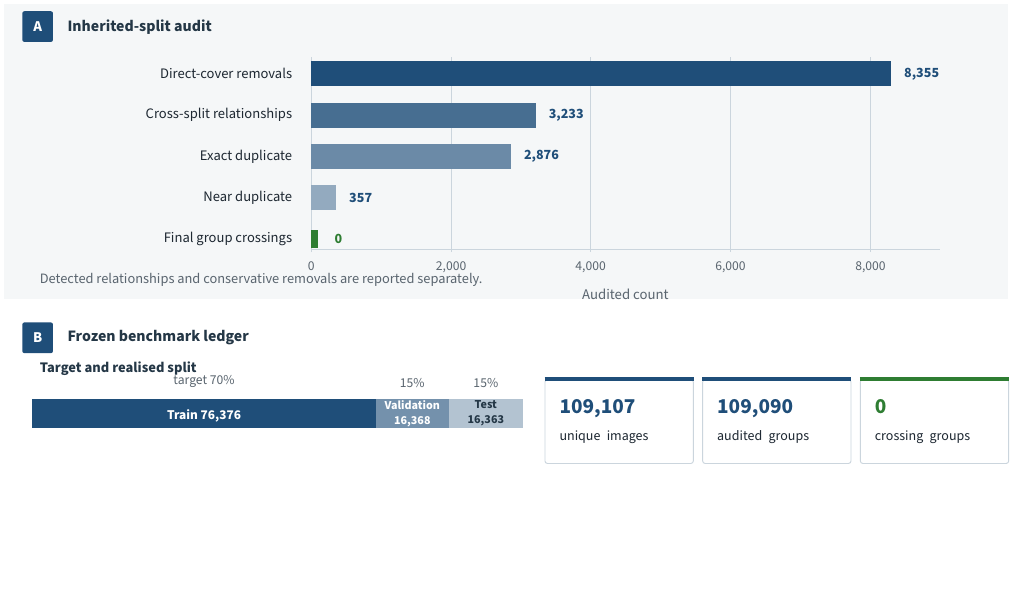}
\caption{Historical contamination and the frozen replacement.}
\label{fig:benchmark-reconstruction}
\FigureNote{Counts refer to audited trusted relationships and groups; undetected duplicate relationships remain possible.}
\end{figure}

A model-readiness stage then prioritised 180 images for human review. Manual review retained 137 images, deleted 43 black-screen or severely distorted images, and relabelled none. Model disagreement was never treated as automatic proof of annotation error. The surviving archive does not preserve reviewer count, blinding, independent duplicate review, or inter-rater agreement, so no such protocol claim is made. Figure~\ref{fig:benchmark-reconstruction} summarises the historical contamination and frozen replacement.

\subsection{Leakage-group-aware split and final freeze}\label{leakage-group-aware-split-and-final-freeze}

Surviving files were assigned by leakage group, not independently by image, against 70/15/15 target shares. The frozen archive records seed 42, five group-safe cross-validation folds over train plus validation, and the resulting assignment of 76,376 training, 16,368 validation, and 16,363 test images across 120 classes. The final build contains 109,107 unique SHA-256 identities and 109,090 audited trusted leakage groups; none of those audited groups spans train, validation, and test. The later manual-review freeze was removal-only: 43 rejected images were deleted without moving any retained row or re-running the split. The archive preserves the final row/group assignment and fingerprints but not a standalone implementation or complete tie-breaking trace of the earlier v5 assignment heuristic; exact historical split regeneration from aggregate summaries is therefore not claimed. Table~\ref{tab:benchmark-ledger} records the resulting audit and freeze ledger.

\begin{table}[H]
\caption{Dataset audit and frozen benchmark ledger.}\label{tab:benchmark-ledger}
\centering
\TableText
\begin{tabularx}{0.88\linewidth}{@{}X>{\raggedleft\arraybackslash}p{0.29\linewidth}@{}}
\rowcolor{BenchmarkBlue}
\color{white}\bfseries Property & \color{white}\bfseries Value \\
\rowcolor{SurfaceGrey} Audited source images & 117,546 \\
Final images & 109,107 \\
\rowcolor{SurfaceGrey} Operational classes & 120 \\
Train / validation / test & 76,376 / 16,368 / 16,363 \\
\rowcolor{SurfaceGrey} Historical duplicate relationships & 8,672 \\
Historical cross-split duplicate relationships & 3,233 \\
\rowcolor{SurfaceGrey} Direct-cover duplicate removals & 8,355 \\
Final leakage groups & 109,090 \\
\rowcolor{PassGreen!8} Leakage groups crossing final splits & \textbf{0} \\
Largest:smallest class ratio & 151.7$\times$ \\
\end{tabularx}
\end{table}

The class distribution remained severely imbalanced. The smallest total class contained 28 images and the largest 4,248. Fifteen classes had fewer than 20 test examples. Consequently, one error can move a four-example class by 25 percentage points. This is why class support accompanies per-class metrics and why the paper avoids presenting rare-class percentages as equally stable estimates.

\subsection{Model-readiness diagnosis and residual risk}\label{model-readiness-diagnosis-and-residual-risk}

Cleaning answered whether known image families crossed the split. It did not answer what a model might learn instead of disease morphology. Before full fine-tuning, frozen DINOv3 ViT-S/16 embeddings were used for a linear probe, exact-neighbour analysis, centroid comparisons, and review prioritisation. The test linear probe reached 85.39\% accuracy but only 68.70\% macro-F1. The 16.69-point gap exposed the long tail before task-specific adaptation.

\begin{figure}[H]
\centering
\includegraphics[width=\linewidth]{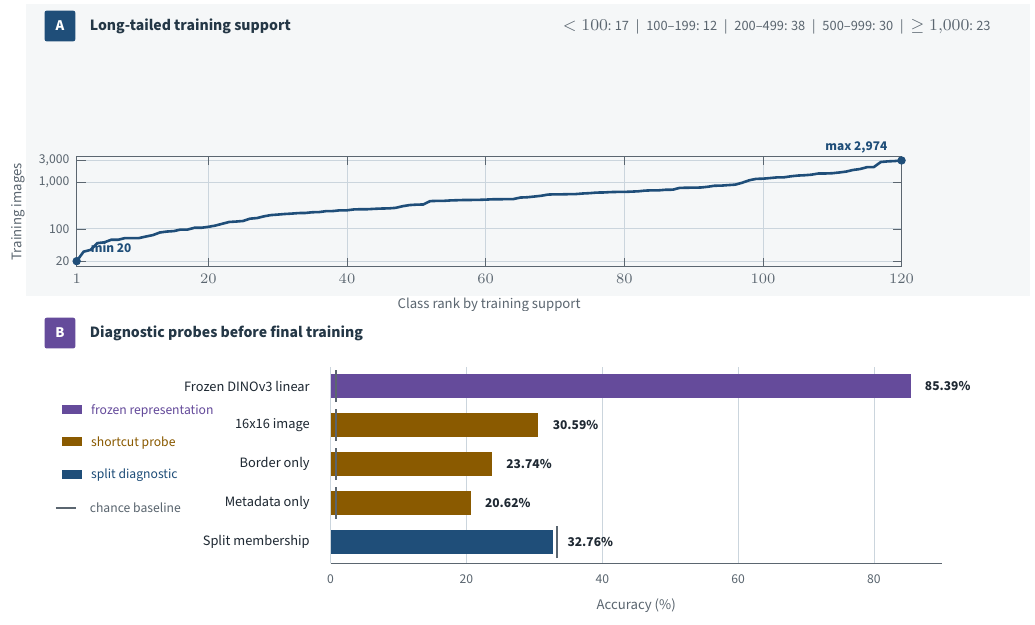}
\caption{Long-tail training support and pre-training diagnostic probes.}
\label{fig:task-diagnostics}
\FigureNote{Shortcut-probe performance above chance motivates the internal-validity boundary.}
\end{figure}

Separate probes measured label information in acquisition-associated features. Technical metadata reached 20.62\% accuracy, 16×16 coarse imagery reached 30.59\%, and border/framing features reached 23.74\%, compared with 0.83\% chance across 120 classes. By contrast, a split-membership classifier reached 32.76\% against 33.33\% chance. The combination is revealing: source and composition cues are label-predictive, yet the final partitions are well mixed in the frozen representation. Internal test performance can therefore be valid for the aggregate distribution while remaining optimistic for new sources. Figure~\ref{fig:task-diagnostics} places the support distribution beside the diagnostic probes.

\FloatBarrier
\clearpage
\section{Model and Training Methodology}\label{model-and-training-methodology}

\subsection{DINOv3 ConvNeXt-Tiny reference}\label{dinov3-convnext-tiny-reference}

The reference model was \Artifact{facebook/dinov3-convnext-tiny-pretrain-lvd1689m} with a 120-way classifier at 256×256 input resolution. The entire backbone was fine-tuned. Validation macro-F1 selected the checkpoint because the diagnostic probe had already shown that accuracy and class-balanced performance could diverge. The completed configuration is recorded below. Optimiser betas and epsilon were not explicitly preserved in the reviewed notebook and are not reconstructed by assumption.

\begingroup
\TableText
\setlength{\tabcolsep}{5.4pt}
\begin{longtable}{@{}>{\RaggedRight\arraybackslash}p{0.285\linewidth}>{\RaggedRight\arraybackslash}p{0.675\linewidth}@{}}
\caption{Completed DINOv3 reference-training configuration.}\label{tab:reference-config}\\
\rowcolor{BenchmarkBlue}
\color{white}\bfseries Field & \color{white}\bfseries Completed reference configuration \\
\endfirsthead
\multicolumn{2}{@{}l}{\sffamily\fontsize{8}{9.8}\selectfont\color{MutedSlate}\textbf{Table \thetable\ --- continued}}\\[4pt]
\rowcolor{BenchmarkBlue}
\color{white}\bfseries Field & \color{white}\bfseries Completed reference configuration \\
\endhead
\rowcolor{SurfaceGrey}
\textbf{Backbone} & \Artifact{facebook/dinov3-convnext-tiny-pretrain-lvd1689m} \\
\textbf{Input / classes} & 256×256 / 120 \\
\rowcolor{WarningAmber!8}
\textbf{Optimiser} & AdamW (betas and epsilon not explicitly preserved) \\
\textbf{Backbone / head learning rate} & \Code{5e-05} / \Code{3e-04} \\
\rowcolor{SurfaceGrey}
\textbf{Weight decay} & 0.05 \\
\textbf{Epochs / best epoch} & 40 / 14 \\
\rowcolor{SurfaceGrey}
\textbf{Warm-up / minimum LR fraction} & 0.05 / 0.02 \\
\textbf{Scheduler} & cosine with warm-up \\
\rowcolor{SurfaceGrey}
\textbf{Gradient clip norm} & 1.0 \\
\textbf{Micro-batch / effective batch} & 32 / 64 \\
\rowcolor{SurfaceGrey}
\textbf{AMP / EMA} & \Code{True} / \Code{True} (decay 0.9998) \\
\textbf{Label smoothing} & 0.02 \\
\rowcolor{SurfaceGrey}
\textbf{Seed} & \Code{42} \\
\textbf{Selection rule} & validation macro-F1 \\
\rowcolor{SurfaceGrey}
\textbf{Geometric augmentation} & RRC \Code{p=0.78}, scale 0.70--1.00, ratio 0.75--1.33; flip \Code{p=0.5}; affine \Code{p=0.4}, ±12°, translation 0.05, scale 0.9--1.1 \\
\textbf{Photometric / codec augmentation} & brightness 0.12, contrast 0.12, saturation 0.08, hue 0.01; JPEG \Code{p=0.10}, quality 65--100; blur/noise \Code{p=0.05} each \\
\rowcolor{SurfaceGrey}
\textbf{Evaluation preprocessing} & EXIF orientation, deterministic RGB conversion, aspect-preserving mean-colour square padding, bicubic resize, normalization \\
\end{longtable}
\endgroup

The augmentation policy was intentionally conservative for disease morphology. It combined controlled crop, affine, illumination, JPEG, blur, and noise variation without making grayscale, solarisation, large hue shifts, or extreme crops mandatory. Deterministic evaluation used EXIF-aware RGB conversion, aspect-preserving square padding, bicubic resize, and the frozen normalisation contract. The selected reference checkpoint was the human-readable epoch 14 and is identified by SHA-256 in Appendix C.

\subsection{Deterministic and resumable training engine}\label{deterministic-and-resumable-training-engine}

Training ran under Kaggle session limits, so resumption was designed as part of experimental identity. Startup checks validated dataset and class-map fingerprints, split counts, duplicate identities, and leakage groups. Augmentation views were derived from the seed, epoch, and frozen row identifier. Epoch shuffling was deterministic, and mid-epoch checkpoints were written only at optimiser-step boundaries. A resumable state included model, optimiser, scheduler, AMP scaler, EMA, random-number generators, epoch, sample offset, global step, and configuration fingerprint. Atomic temporary-to-final replacement prevented partially written checkpoints from being treated as valid states.

These controls do not raise accuracy by themselves. They prevent an interrupted session from silently becoming a different experiment. Ten AMP-skipped updates were recorded; optimiser-dependent state was not advanced when those updates were skipped. The final evidence also includes AMP-versus-FP32 and batch-size parity checks, allowing numerical implementation effects to be separated from model-selection effects.

\subsection{Locked evaluation and uncertainty}\label{locked-evaluation-and-uncertainty}

The test loader was disabled by default and required explicit authorization after the reference configuration, class map, preprocessing contract, and evaluation plan were frozen. Validation and test outputs preserved stable row identifiers, labels, logits, class order, and model/data fingerprints. The metric suite included accuracy, balanced accuracy, macro- and weighted F1, top-3 and top-5 accuracy, negative log-likelihood, Brier score, 15-bin expected calibration error, support slices, and complete per-class reports.

A class-stratified bootstrap over the fixed test rows produced uncertainty intervals for the reference estimate. Within each true class, rows were sampled with replacement at the original class size; paired comparisons applied the same sampled indices to both states. The reported intervals are percentile intervals from the 2.5th and 97.5th quantiles. For the reference-to-PTE analysis, 2,000 replicates used seed 20260821, and calibration temperatures and selective-prediction thresholds were not re-estimated inside replicates. This bootstrap describes sensitivity to the composition of the locked test sample; it does not measure retraining variability. The completed lineage contains one archived training seed for the reference and one for the compact model.

\begin{reportingnotebox}
\textbf{Numeric reporting.} Headline prose, stat cards, and ordinary descriptive tables report percentages to two decimal places. Additional recorded precision is retained where it is necessary to distinguish PTQ candidates, model-state transitions, paired intervals, row-level confidences, or exact artifact identity. Appendix statistical tables preserve the released full precision.
\end{reportingnotebox}

\subsection{Compact MobileNetV4 derivative}\label{compact-mobilenetv4-derivative}

The compact state used \Artifact{mobilenetv4\_conv\_medium.e500\_r256\_in1k} from timm \Code{1.0.26}, a 256×256 input, ImageNet normalisation, bicubic interpolation, and a 120-way linear classifier. Direct counting from the archived selected state gives 8,588,232 parameter tensors when batch-normalisation running statistics and counters are excluded. Exact MACs, activation memory, and peak resident memory were not measured for the frozen 120-class graph and are not inferred from a generic ImageNet model card. The 30-epoch history records total loss, hard-label loss, teacher-logit loss, feature-transfer loss, teacher reliability, and teacher top-1 correctness. The selected state was the raw best checkpoint. Train-only batch-normalisation recalibration and delayed-EMA-plus-BN alternatives did not improve the selected validation result.

\begin{evidenceboundarybox}
The archive identifies the exact selected state and configuration fingerprint, but it does not preserve a complete human-readable record of the mobile optimiser, learning-rate schedule, distillation temperature, or objective coefficients. We therefore report the components that are directly recorded and do not claim exact recipe-level reproducibility for the historical mobile run. We also do not attribute the student's performance causally to distillation: a matched direct MobileNetV4 run is absent. The supported statement is descriptive---the completed teacher-guided lineage reached the reported metrics.
\end{evidenceboundarybox}

\subsection{State identity and comparison design}\label{state-identity-and-comparison-design}

Four model states are kept separate throughout the analysis: the DINOv3 reference, the selected float MobileNetV4 state, the converted INT8 graph, and the serialised PTE. The float and quantised states operate on the same locked examples with the same class ordering. Pairwise comparisons therefore report both ground-truth metrics and top-1 agreement. Agreement is not treated as accuracy: two states can agree on the same wrong class. Every state-specific result is attached to raw predictions or logits, and the final artifact is attached to an exact SHA-256 digest.

\FloatBarrier
\clearpage
\section{Quantisation and Runtime Methodology}\label{quantisation-and-runtime-methodology}

\subsection{Validation-only PTQ selection}\label{validation-only-ptq-selection}

Post-training quantisation was treated as a model-selection problem rather than a file-conversion formality. Three predeclared XNNPACK-compatible candidates were evaluated on the full validation split before test access. The selection rule was pass-first and macro-F1-first, with a small tie preference defined in the frozen recipe. The candidates included dynamic activation quantisation with per-channel weights and two calibrated static alternatives. Only the selected recipe was carried into the locked test path.

\begin{table}[H]
\centering
\caption{Validation-only post-training quantisation candidates.}
\label{tab:ptq-candidates}
\begingroup
\sffamily\fontsize{8}{10}\selectfont
\setlength{\tabcolsep}{3.4pt}
\renewcommand{\arraystretch}{1.38}
\begin{tabularx}{\textwidth}{@{}>{\RaggedRight\arraybackslash}p{0.082\textwidth}>{\RaggedRight\arraybackslash}X
S[table-format=2.4,group-digits=false]S[table-format=2.4,group-digits=false]S[table-format=2.4,group-digits=false]S[table-format=2.4,group-digits=false]S[table-format=4.0,group-digits=integer]>{\centering\arraybackslash}p{0.074\textwidth}@{}}
\rowcolor{BenchmarkBlue}
\color{white}\bfseries Candidate &
\color{white}\bfseries Scheme &
\multicolumn{1}{c}{\color{white}\bfseries \shortstack{Val.\\accuracy\\(\%)}} &
\multicolumn{1}{c}{\color{white}\bfseries \shortstack{Val. balanced\\accuracy (\%)}} &
\multicolumn{1}{c}{\color{white}\bfseries \shortstack{Val.\\macro-F1 (\%)}} &
\multicolumn{1}{c}{\color{white}\bfseries \shortstack{Agreement\\with float (\%)}} &
\multicolumn{1}{c}{\color{white}\bfseries \shortstack{Calibration\\rows}} &
\color{white}\bfseries Gate \\
\rowcolor{BenchmarkBlue!8}
\textbf{Dynamic} & Dynamic activations; per-channel weights & 98.4726 & 96.5189 & 96.4882 & 99.9389 & 0 & \StatusPass \\
\rowcolor{SurfaceGrey}
\textbf{Cal16} & Static; 2,880 calibration rows & 98.4299 & 96.4088 & 96.3458 & 99.6884 & 2880 & \StatusPass \\
\rowcolor{FailureRed!5}
\textbf{Cal8} & Static; 1,440 calibration rows & 98.4421 & 96.3787 & 96.3461 & 99.7006 & 1440 & \StatusFail \\
\end{tabularx}
\endgroup
\ObjectNote{Repository identifiers: \Code{pc\_dynamic\_full}, \Code{pc\_full\_cal16}, and \Code{pc\_full\_cal8}.}
\end{table}

The dynamic candidate achieved the strongest validation macro-F1, class-balanced result, and float agreement, while requiring no calibration rows. The 2,880-row static candidate passed the quality gate but was weaker. The 1,440-row static candidate failed the predeclared gate. These results do not establish that dynamic quantisation is universally superior; they justify the selected path for this model, backend, and validation distribution.

\subsection{Converted-graph evaluation}\label{converted-graph-evaluation}

The selected converted graph was evaluated on the same locked test rows as the float state. Its accuracy changed by −0.006 percentage points, balanced accuracy by −0.048 points, and macro-F1 by −0.022 points. Float and converted INT8 predictions agreed on 99.951\% of rows. A paired class-stratified bootstrap placed the converted-minus-float macro-F1 change between −0.097 and +0.024 percentage points at 95\% confidence. The interval includes zero, supporting the practical conclusion that the selected conversion did not produce a clear aggregate class-balanced loss on the fixed test.

\subsection{Serialisation and direct PTE execution}\label{serialisation-and-direct-pte-execution}

The converted graph was lowered to ExecuTorch with the XNNPACK backend and serialised as \Code{CropCop\_Mobile\_INT8\_}\allowbreak\Code{XNNPACK\_PRODUCTION\_v1.pte}. The exact artifact contains 23,696,352 bytes, or 22.5986 MiB, and is identified by SHA-256 \Code{7c70d0f307f0d9578310600913cb8ff1}\allowbreak\Code{71b294ae4de0813f7cd1d6a53628bdf1}. It is 31.92\% smaller than the selected 34,805,405-byte pre-quant state file. That comparison is between two different containers, so it is reported as an artifact-size reduction rather than a theoretical fourfold INT8 compression claim.

Direct PTE execution generated a new prediction record on all 16,363 locked rows. The PTE is therefore not assigned the converted graph's metrics by inheritance. Its accuracy, macro-F1, agreement, and disagreements are measured from the serialised program itself.

\subsection{Runtime environment and quantisation coverage}\label{runtime-environment-and-quantisation-coverage}

\begin{table}[H]
\centering
\caption{Archived runtime environment and quantisation coverage.}
\label{tab:runtime-summary}
\begingroup
\TableText
\setlength{\tabcolsep}{5.4pt}
\renewcommand{\arraystretch}{1.28}
\begin{tabularx}{\linewidth}{@{}>{\RaggedRight\arraybackslash}p{0.255\linewidth}>{\RaggedRight\arraybackslash}X@{}}
\rowcolor{BenchmarkBlue}
\color{white}\bfseries Evidence domain & \color{white}\bfseries Archived value \\
\rowcolor{BenchmarkBlue!8}
\textbf{Execution context} & Kaggle-host CPU software-runtime evaluation; not Android device evidence \\
\rowcolor{SurfaceGrey}
\textbf{Software stack} & Python \Code{3.12.13}; PyTorch / torchvision \Code{2.12.1 / 0.27.1}; ExecuTorch / torchao \Code{1.3.1 / 0.17.0}; timm / FlatBuffers \Code{1.0.26 / 25.12.19} \\
\textbf{Backend / input} & XNNPACK / \Code{[1, 3, 256, 256]} \\
\rowcolor{SurfaceGrey}
\textbf{Quantisation contract} & Requested scope \Code{full}; dynamic activation / per-channel weights; explicit module exclusions \Code{[]} \\
\rowcolor{PassGreen!7}
\textbf{Coverage evidence} & \StatusPass\quad Lowered backend modules / delegate tokens: \Code{1 / 1} \\
\rowcolor{WarningAmber!7}
\textbf{Unarchived host details} & CPU model, thread count, and OS build \\
\rowcolor{WarningAmber!7}
\textbf{Physical Android evaluation} & Not evaluated \\
\end{tabularx}
\endgroup
\ObjectNote{The complete field-by-field environment register is retained for Appendix D in the full manuscript.}
\end{table}

The archived graph records a full requested quantisation scope, no explicit module exclusions, a passed project coverage gate, and one lowered XNNPACK backend module/delegate token. These records support the intended backend path, but they do not provide an operator-level delegation dump: total delegated nodes, portable-fallback nodes, and partition boundaries were not archived. The paper therefore does not claim complete XNNPACK operator coverage. A host-only timing record exists, yet the CPU model, thread count, and operating-system build were not archived, so those latencies are excluded from the main performance claims. These omissions prevent host-latency reproduction but do not alter the reported INT8 prediction-fidelity comparison, which is computed from archived converted-graph and PTE outputs on the same locked rows. Android latency, memory, energy, delegate fallback, and thermal behaviour remain unmeasured.

\FloatBarrier
\clearpage
\section{Results}\label{results}

\subsection{End-to-end model-state performance}\label{end-to-end-model-state-performance}

The primary result is the retention of class-balanced performance across identified model states. The DINOv3 reference achieved 98.51\% accuracy and 96.87\% macro-F1. The float MobileNetV4 state finished 0.05 accuracy points and 0.60 macro-F1 points lower. Conversion to INT8 produced little further movement. Direct execution of the final PTE returned to the float state's accuracy and finished 0.04 macro-F1 points below it.

\begin{table}[H]
\centering
\caption{Performance across identified model states on the locked internal test.}
\label{tab:model-state-performance}
\begingroup
\sffamily\fontsize{7.9}{9.8}\selectfont
\setlength{\tabcolsep}{3.4pt}
\renewcommand{\arraystretch}{1.36}
\begin{tabularx}{\textwidth}{@{}>{\RaggedRight\arraybackslash}p{0.105\textwidth}>{\RaggedRight\arraybackslash}X
S[table-format=2.4,group-digits=false]S[table-format=2.4,group-digits=false]S[table-format=2.4,group-digits=false]S[table-format=2.4,group-digits=false]S[table-format=2.4,group-digits=false]S[table-format=3.0]@{}}
\rowcolor{BenchmarkBlue}
\color{white}\bfseries State &
\color{white}\bfseries Architecture / object &
\multicolumn{1}{c}{\color{white}\bfseries \shortstack{Accuracy\\(\%)}} &
\multicolumn{1}{c}{\color{white}\bfseries \shortstack{Balanced\\accuracy (\%)}} &
\multicolumn{1}{c}{\color{white}\bfseries \shortstack{Macro-F1\\(\%)}} &
\multicolumn{1}{c}{\color{white}\bfseries \shortstack{Top-3\\(\%)}} &
\multicolumn{1}{c}{\color{white}\bfseries \shortstack{Top-5\\(\%)}} &
\multicolumn{1}{c}{\color{white}\bfseries Errors} \\
\rowcolor{BenchmarkBlue!8}
\textbf{Reference} & DINOv3 ConvNeXt-Tiny & 98.5088 & 96.6836 & 96.8700 & 99.7250 & 99.8533 & 244 \\
\rowcolor{SurfaceGrey}
\textbf{Float student} & MobileNetV4 Conv-Medium & 98.4599 & 96.2435 & 96.2710 & 99.7494 & 99.8594 & 252 \\
\textbf{Converted INT8} & XNNPACK-compatible converted graph & 98.4538 & 96.1957 & 96.2492 & 99.7372 & 99.8533 & 253 \\
\rowcolor{SurfaceGrey}
\textbf{PTE runtime} & Serialised ExecuTorch/XNNPACK program & 98.4599 & 96.2017 & 96.2267 & 99.7433 & 99.8717 & 252 \\
\end{tabularx}
\endgroup
\end{table}

\begin{figure}[t]
\centering
\includegraphics[width=\linewidth]{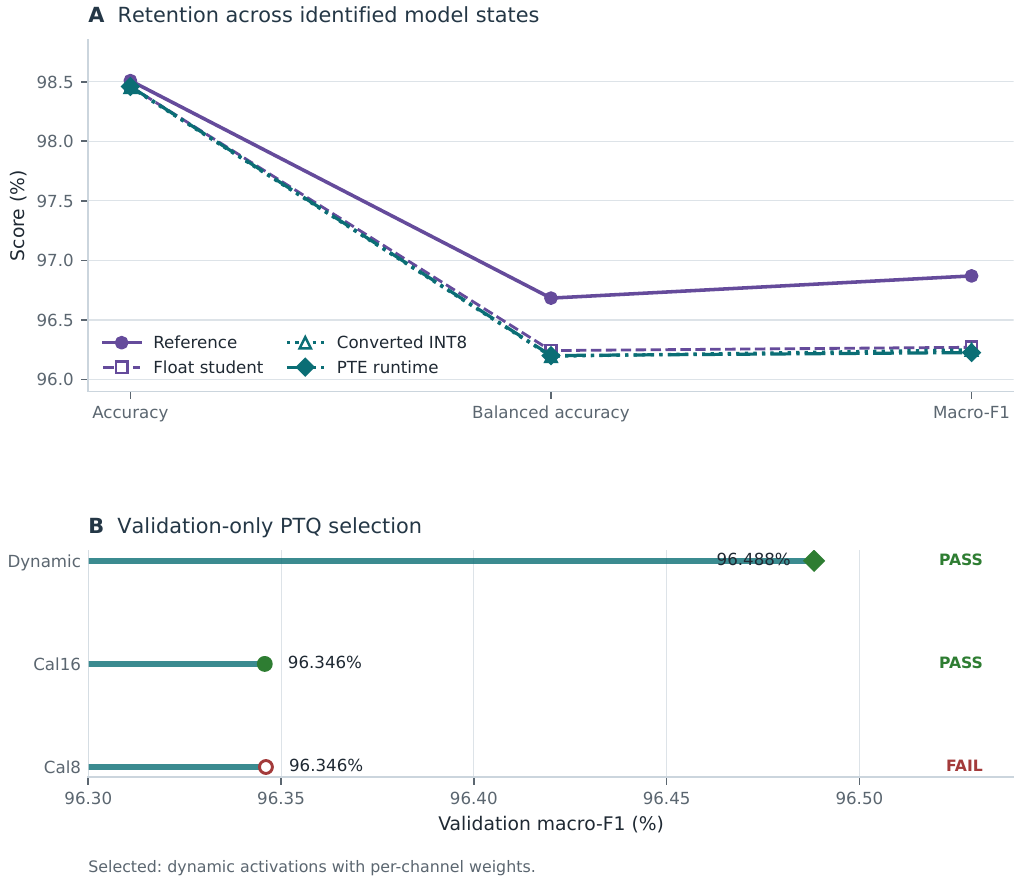}
\caption{Accuracy, balanced accuracy, and macro-F1 across the four identified model states, with validation-only PTQ candidate results.}
\label{fig:4}
\end{figure}

The table should not be read as an architecture tournament. The reference and compact states differ in architecture, training procedure, and role. It answers a narrower engineering question: how much of the strong reference result survived the completed compact and runtime path? On aggregate accuracy, nearly all of it survived. Macro-F1 reveals a small but measurable redistribution across classes. The four-state and PTQ results are visualised in Figure~\ref{fig:4}.

\subsection{Benchmark reconstruction and task diagnosis}\label{benchmark-reconstruction-and-task-diagnosis}

The benchmark result precedes the model result logically even though it follows it in presentation. The historical split was abandoned because 3,233 confirmed duplicate relationships crossed its boundaries. The frozen replacement contains 109,107 unique images and zero crossings among the audited trusted leakage groups. This removes a known source of train--test family overlap, but it does not remove source bias or ontology ambiguity.

The diagnostic probes clarify the remaining problem. A frozen DINOv3 representation reached 85.39\% accuracy and 68.70\% macro-F1, showing that much of the label structure was visible before fine-tuning while minority performance lagged. Metadata, 16×16 imagery, and border probes achieved 20.62\%, 30.59\%, and 23.74\% accuracy, all far above chance. Split membership remained at chance. The final split is internally coherent, yet it shares label-correlated acquisition cues across partitions. This is why the paper uses the phrase \emph{leakage-controlled internal performance} rather than \emph{field generalisation}.

\subsection{Paired reference-to-PTE analysis}\label{paired-reference-to-pte-analysis}

Point estimates alone understate the structure of the reference-to-runtime change. Across the same 16,363 rows, the reference and PTE differed on 240 top-1 predictions. The reference was uniquely correct on 105 of those rows and the PTE uniquely correct on 97; 38 rows were wrong under both states but assigned different labels. The PTE made 252 errors, eight more than the reference. An exact two-sided McNemar test on the correctness indicators gave \emph{p} = 0.622 \citep{mcnemar1947sampling}, so the aggregate accuracy difference is not distinguishable in that paired test.

Class-balanced behaviour tells a sharper story. In 2,000 class-stratified paired bootstrap replicates, the PTE's accuracy change had a 95\% interval spanning zero, as did the balanced-accuracy change. The macro-F1 change was −0.64 percentage points with a 95\% interval of −1.24 to −0.05 points. An exploratory post hoc slice defined mechanically by labels containing the literal substring \Code{fruit} showed a 6.97-point mean-recall decline, with a percentile interval of −12.41 to −1.71 points. Its exact membership is released with the evidence package; because the slice was examined during error analysis rather than preregistered, it is descriptive and hypothesis-generating. The overall result is therefore not ``no loss.'' It is near-parity in aggregate correctness with a modest, concentrated reduction in class-balanced performance.

\begin{table}[H]
\centering
\caption{Paired reference-to-PTE changes with class-stratified percentile intervals.}
\label{tab:paired-summary}
\begingroup
\TableText
\setlength{\tabcolsep}{6.2pt}
\renewcommand{\arraystretch}{1.34}
\begin{tabularx}{\textwidth}{@{}>{\RaggedRight\arraybackslash}X
S[table-format=2.3,group-digits=false]S[table-format=2.3,group-digits=false]>{\raggedleft\arraybackslash}p{0.17\textwidth}>{\raggedleft\arraybackslash}p{0.235\textwidth}@{}}
\rowcolor{BenchmarkBlue}
\color{white}\bfseries Metric &
\multicolumn{1}{c}{\color{white}\bfseries \shortstack{Reference\\(\%)}} &
\multicolumn{1}{c}{\color{white}\bfseries \shortstack{PTE\\(\%)}} &
\color{white}\bfseries \shortstack{PTE $-$ reference\\(pp)} &
\color{white}\bfseries \shortstack{Paired 95\% interval\\(pp)} \\
\rowcolor{BenchmarkBlue!8}
\textbf{Accuracy} & 98.509 & 98.460 & $-0.049$ & $[-0.214, +0.116]$ \\
\rowcolor{SurfaceGrey}
\textbf{Balanced accuracy} & 96.684 & 96.202 & $-0.482$ & $[-1.097, +0.113]$ \\
\textbf{Macro-F1} & 96.870 & 96.227 & $-0.643$ & $[-1.239, -0.047]$ \\
\rowcolor{SurfaceGrey}
\textbf{Fruit-category mean recall} & 85.936 & 78.971 & $-6.965$ & $[-12.409, -1.711]$ \\
\end{tabularx}
\endgroup
\ObjectNote{Full-precision estimates and support-bin recall intervals are reported in Appendix~\ref{reference-to-pte-paired-bootstrap}.}
\end{table}

\subsection{Long-tail retention and class-level redistribution}\label{long-tail-retention-and-class-level-redistribution}

Disjoint training-support bins show where the compact runtime paid the largest cost. Classes with fewer than 100 training examples lost 2.56 macro-F1 points on average, and the 100--199 bin lost 2.78 points. The three higher-support bins changed by less than one tenth of a point in magnitude, except that the ≥1,000 bin improved slightly. These are descriptive averages over heterogeneous classes; support alone does not determine difficulty.

\begin{table}[H]
\centering
\caption{Performance retention by disjoint training-support bin.}
\label{tab:support-bin-retention}
\begingroup
\sffamily\fontsize{8}{9.9}\selectfont
\setlength{\tabcolsep}{3.8pt}
\renewcommand{\arraystretch}{1.35}
\begin{tabularx}{\textwidth}{@{}>{\RaggedRight\arraybackslash}X
>{\raggedleft\arraybackslash}p{0.064\textwidth}
>{\raggedleft\arraybackslash}p{0.103\textwidth}
>{\raggedleft\arraybackslash}p{0.097\textwidth}
>{\raggedleft\arraybackslash}p{0.078\textwidth}
>{\raggedleft\arraybackslash}p{0.103\textwidth}
>{\raggedleft\arraybackslash}p{0.097\textwidth}
>{\raggedleft\arraybackslash}p{0.078\textwidth}@{}}
\rowcolor{BenchmarkBlue}
\color{white}\bfseries \shortstack{Training\\support} &
\color{white}\bfseries Classes &
\color{white}\bfseries \shortstack{Reference\\macro-F1 (\%)} &
\color{white}\bfseries \shortstack{PTE\\macro-F1 (\%)} &
\color{white}\bfseries \shortstack{Change\\(pp)} &
\color{white}\bfseries \shortstack{Reference\\recall (\%)} &
\color{white}\bfseries \shortstack{PTE\\recall (\%)} &
\color{white}\bfseries \shortstack{Change\\(pp)} \\
\rowcolor{BenchmarkBlue!8}
$<100$ & 17 & 91.17 & 88.60 & $-2.56$ & 89.37 & 87.27 & $-2.10$ \\
\rowcolor{SurfaceGrey}
100--199 & 12 & 92.84 & 90.05 & $-2.78$ & 93.09 & 90.27 & $-2.82$ \\
200--499 & 38 & 97.91 & 97.85 & $-0.06$ & 98.04 & 98.34 & $+0.30$ \\
\rowcolor{SurfaceGrey}
500--999 & 30 & 99.07 & 98.98 & $-0.09$ & 99.10 & 99.01 & $-0.09$ \\
$\geq1000$ & 23 & 98.60 & 98.81 & $+0.21$ & 98.58 & 98.71 & $+0.13$ \\
\end{tabularx}
\endgroup
\end{table}

The per-class deltas reinforce that warning, but their ranking is descriptive: several classes have only 7--30 test examples, so small changes in TP, FP, or FN can move F1 sharply. Exact reconstructed TP/FP/FN counts are released beside the delta table. \nolinkurl{banana_disease_fruit} fell from 0.7778 to 0.6000 F1 on 11 test images. Broad fruit-condition categories account for several of the largest declines, including \nolinkurl{mango_disease_fruit}, \nolinkurl{guava_healthy_fruit}, and \nolinkurl{papaya_fruit_healthy}. By contrast, \nolinkurl{plant_healthy} improved from 0.7609 to 0.8824 F1. Compression did not merely lower every score by a fixed amount; it altered a small set of class boundaries.

\begin{table}[H]
\centering
\caption{Largest class-level F1 declines.}
\label{tab:largest-class-declines}
\begingroup
\sffamily\fontsize{8}{9.8}\selectfont
\setlength{\tabcolsep}{5.2pt}
\renewcommand{\arraystretch}{1.27}
\begin{tabularx}{\textwidth}{@{}>{\RaggedRight\arraybackslash}X
S[table-format=3.0]S[table-format=2.0]S[table-format=1.4,group-digits=false]S[table-format=1.4,group-digits=false]S[table-format=-1.4,group-digits=false]@{}}
\rowcolor{BenchmarkBlue}
\color{white}\bfseries Class &
\multicolumn{1}{c}{\color{white}\bfseries \shortstack{Train\\support}} &
\multicolumn{1}{c}{\color{white}\bfseries \shortstack{Test\\support}} &
\multicolumn{1}{c}{\color{white}\bfseries \shortstack{Reference F1\\(0--1)}} &
\multicolumn{1}{c}{\color{white}\bfseries \shortstack{PTE F1\\(0--1)}} &
\multicolumn{1}{c}{\color{white}\bfseries \shortstack{$\Delta$ F1\\(0--1)}} \\
\rowcolor{BenchmarkBlue!8}
\nolinkurl{banana_disease_fruit} & 51 & 11 & 0.7778 & 0.6000 & -0.1778 \\
\rowcolor{SurfaceGrey}
\nolinkurl{mango_disease_fruit} & 143 & 30 & 0.8788 & 0.7213 & -0.1575 \\
\nolinkurl{guava_healthy_fruit} & 35 & 7 & 1.0000 & 0.8571 & -0.1429 \\
\rowcolor{SurfaceGrey}
\nolinkurl{guava_disease_fruit} & 110 & 23 & 0.8000 & 0.7234 & -0.0766 \\
\nolinkurl{papaya_fruit_healthy} & 96 & 20 & 0.9231 & 0.8500 & -0.0731 \\
\rowcolor{SurfaceGrey}
\nolinkurl{tomato_mosaic_virus} & 240 & 51 & 0.9703 & 0.9231 & -0.0472 \\
\nolinkurl{guava_healthy_leaf} & 105 & 22 & 1.0000 & 0.9545 & -0.0455 \\
\rowcolor{SurfaceGrey}
\nolinkurl{mango_healthy_fruit} & 227 & 49 & 0.9216 & 0.8824 & -0.0392 \\
\nolinkurl{aloevera_healthy} & 63 & 14 & 1.0000 & 0.9630 & -0.0370 \\
\rowcolor{SurfaceGrey}
\nolinkurl{raspberry_healthy} & 264 & 57 & 0.9655 & 0.9298 & -0.0357 \\
\end{tabularx}
\endgroup
\ObjectNote{The ranking is descriptive: several classes have only 7--30 test examples, so small changes in TP, FP, or FN can move F1 sharply.}
\end{table}

\begin{figure}[H]
\centering
\includegraphics[width=\linewidth]{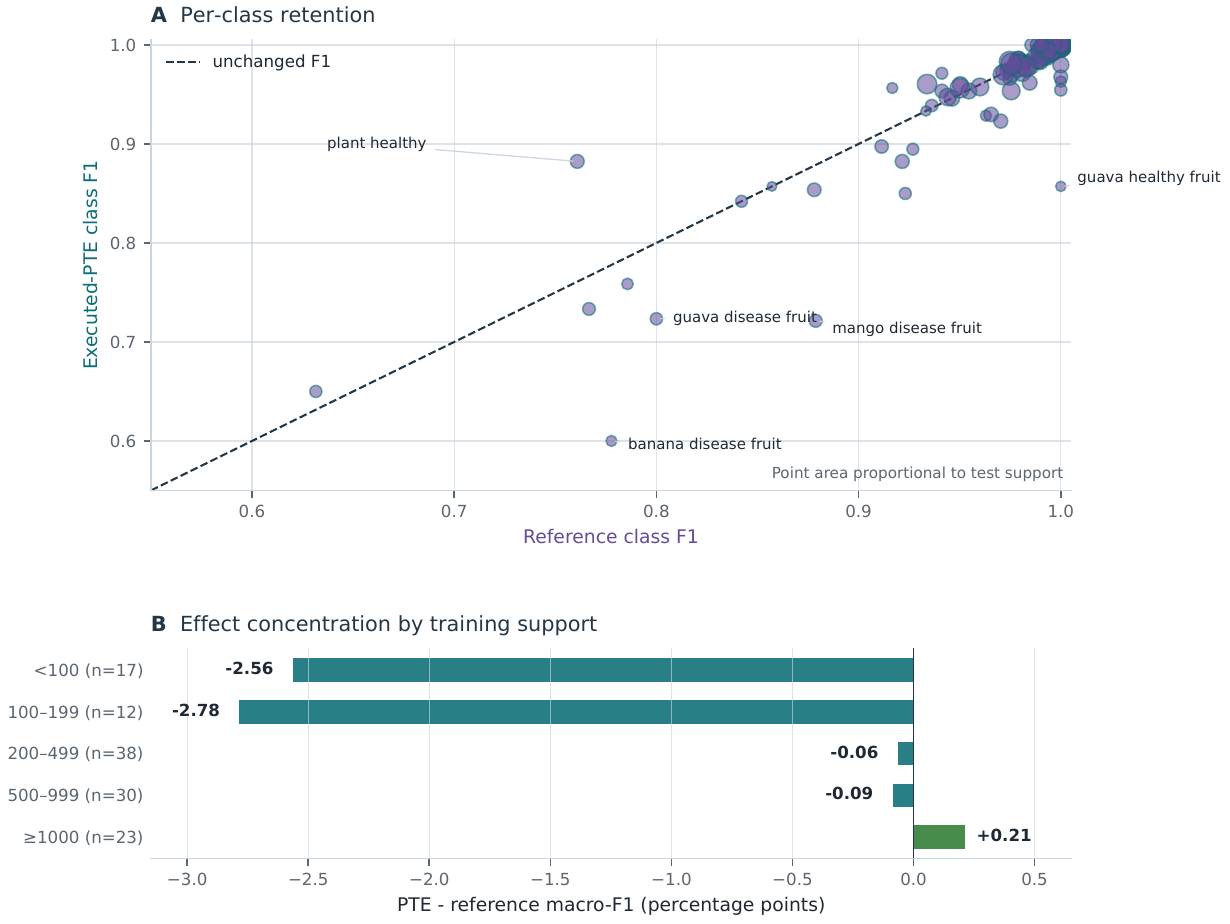}
\caption{Reference-versus-PTE class F1; point area is proportional to test support.}
\label{fig:5}
\FigureNote{Labelled declines are descriptive for small classes.}
\end{figure}

The class-level redistribution is visualised in Figure~\ref{fig:5}.

\subsection{From converted graph to serialised PTE: six changed decisions}\label{from-converted-graph-to-serialised-pte-six-changed-decisions}

Converted INT8 and PTE predictions differed on exactly six of 16,363 rows. Two changes turned correct INT8 predictions into PTE errors, three corrected INT8 errors, and one moved between two wrong classes. As a result, PTE accuracy was one correct prediction higher than converted INT8 accuracy, even though macro-F1 was 0.023 points lower. The difference is possible because each changed prediction affects class precision and recall differently.

None of the six changed decisions was high-margin. Converted confidences ranged from 0.262 to 0.530, and every margin was below 0.13. The serialisation changes therefore occurred near contested decision boundaries rather than among unequivocal predictions. This audit is more informative than agreement alone: 99.963\% agreement becomes a concrete list of six cases that can be inspected and reproduced.

\begin{reportingnotebox}
\textbf{Row-level audit.} The complete six-row converted-INT8-to-PTE disagreement table is retained in Appendix~\ref{converted-int8-to-pte-disagreement-audit}. Figure~\ref{fig:6}B--C preserves the 2 / 3 / 1 transition summary and the margin distribution in the main results narrative.
\end{reportingnotebox}

\subsection{Calibration, selective prediction, and error taxonomy}\label{calibration-selective-prediction-and-error-taxonomy}

Raw PTE ECE15 was 1.58\% and NLL was 0.0799 on the internal test. Calibration numbers describe the frozen distribution, not confidence under domain shift. To examine whether confidence could support review, thresholds were fixed independently from each state's validation outputs at target coverages of 80\%, 90\%, 95\%, and 98\%, then applied once to test outputs. At the 95\% target, the reference covered 94.88\% of test rows at 99.81\% accepted-set accuracy; the PTE covered 94.82\% at 99.67\%. No threshold from this internal analysis is proposed as a field-deployment policy.

\begin{table}[H]
\centering
\caption{Validation-derived selective-prediction operating points.}
\label{tab:selective-prediction}
\begingroup
\TableText
\setlength{\tabcolsep}{6.5pt}
\renewcommand{\arraystretch}{1.34}
\begin{tabularx}{\textwidth}{@{}RRRRR@{}}
\rowcolor{BenchmarkBlue}
\multicolumn{1}{c}{\color{white}\bfseries \shortstack{Validation target\\coverage (\%)}} &
\multicolumn{1}{c}{\color{white}\bfseries \shortstack{Reference test\\coverage (\%)}} &
\multicolumn{1}{c}{\color{white}\bfseries \shortstack{Reference accepted\\accuracy (\%)}} &
\multicolumn{1}{c}{\color{white}\bfseries \shortstack{PTE test\\coverage (\%)}} &
\multicolumn{1}{c}{\color{white}\bfseries \shortstack{PTE accepted\\accuracy (\%)}} \\
\rowcolor{BenchmarkBlue!8}
80 & 79.80 & 99.824 & 79.72 & 99.678 \\
\rowcolor{SurfaceGrey}
90 & 89.79 & 99.816 & 89.90 & 99.681 \\
95 & 94.88 & 99.813 & 94.82 & 99.671 \\
\rowcolor{SurfaceGrey}
98 & 97.99 & 99.451 & 97.90 & 99.407 \\
\end{tabularx}
\endgroup
\end{table}

An explicit threshold sweep further clarifies the phrase \emph{high-confidence error}. For the PTE, 82 wrong predictions had raw softmax confidence above 0.95, 18 exceeded 0.99, and none exceeded 0.999. These counts should not be compared directly with the reference without accounting for its temperature scaling and confidence distribution. Their purpose is to show that even a highly accurate internal model can make confident mistakes.

The descriptive error taxonomy was derived mechanically from operational label strings, not from biological adjudication. Same-crop condition confusions dominated both states. Fruit-condition involvement increased from 47 reference errors to 64 PTE errors, while errors involving the generic \Code{plant\_healthy} label decreased. This pattern agrees with the per-class analysis: the main compression cost is concentrated in broad, visually heterogeneous fruit categories rather than spread evenly across the ontology.

\begin{reportingnotebox}
\textbf{Operational taxonomy.} The complete six-category reference-to-PTE error taxonomy is retained in Appendix~\ref{reference-to-pte-operational-error-taxonomy}. The main-text interpretation preserves the two largest directional changes: fruit-condition involvement increased from 47 to 64 errors, while generic \Code{plant\_healthy} involvement decreased from 22 to 12.
\end{reportingnotebox}

\begin{figure}[H]
\centering
\includegraphics[width=\linewidth]{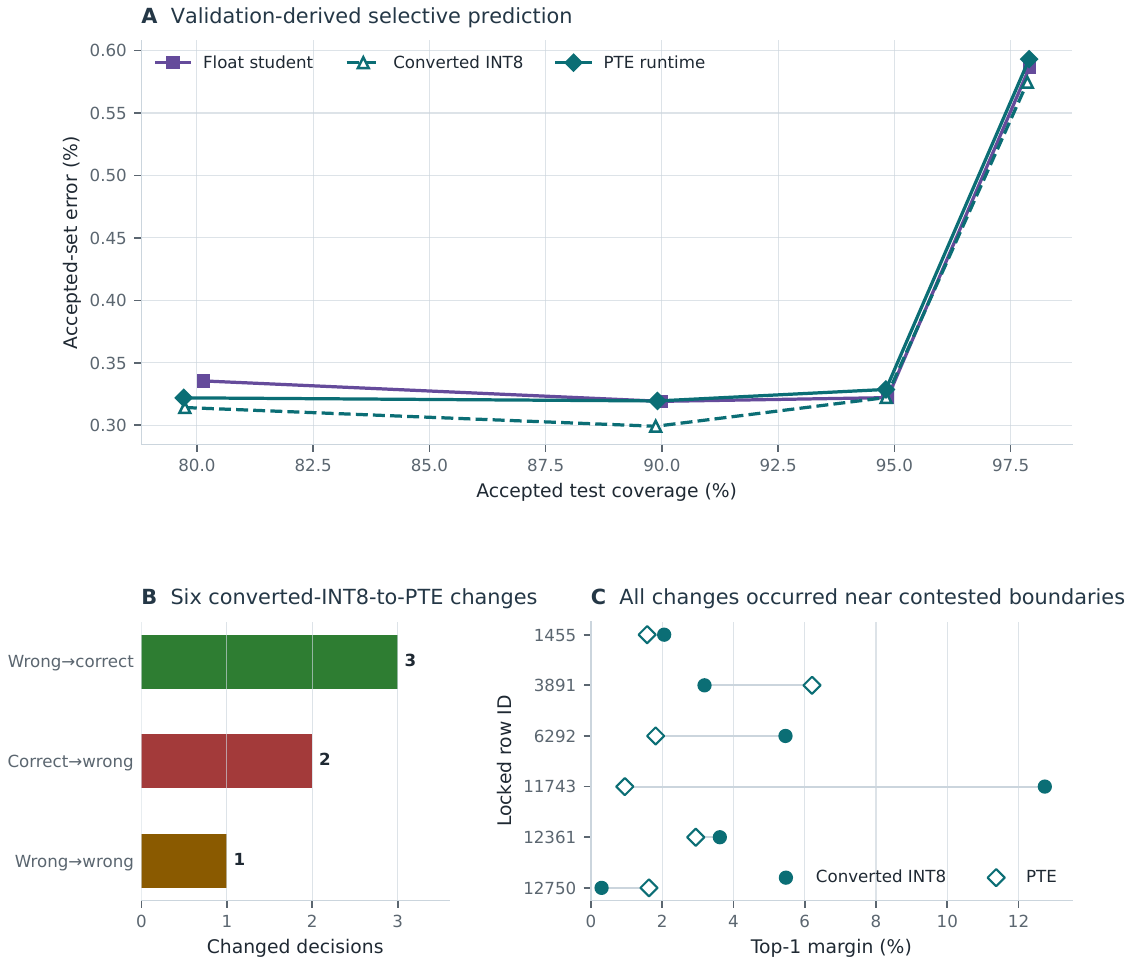}
\caption{Validation-derived selective-prediction operating points and the six converted-INT8-to-PTE decision changes.}
\label{fig:6}
\end{figure}

Figure~\ref{fig:6} summarises the validation-derived operating points and the six converted-INT8-to-PTE changes.

\clearpage
\section{Discussion}\label{discussion}

\subsection{What the paper establishes}\label{what-the-paper-establishes}

CropCop establishes that a broad, long-tailed plant-health model can retain high internal performance across a traceable sequence of benchmark reconstruction, foundation-model adaptation, compact training, quantisation, backend lowering, and direct runtime execution. The final 98.46\% accuracy is not assigned to a project name; it is attached to a particular split, class map, model lineage, prediction record, PTE file, and SHA-256 digest. That traceability is the paper's central contribution. It allows a reviewer to ask where a change occurred and answer with paired rows rather than with speculation.

\subsection{The model result is strong, but not lossless}\label{the-model-result-is-strong-but-not-lossless}

The PTE finished only 0.05 accuracy points below the reference, and the paired correctness comparison did not show a clear aggregate accuracy difference. Macro-F1, however, declined by 0.64 points with a paired interval excluding zero. This is an important distinction. A compact system can preserve nearly every top-1 decision while changing the distribution of those decisions across classes. Reporting only accuracy would erase the main cost of compression; reporting only macro-F1 would obscure how small the row-level change actually was.

\subsection{What the error redistribution teaches}\label{what-the-error-redistribution-teaches}

The weakest changes are not random. Low-support groups and broad fruit-condition labels lose more than high-support groups, while some classes improve. The result points to two distinct constraints. Limited examples make a boundary sensitive to model capacity and training variation. Broad or inconsistent labels impose ambiguity that a larger backbone cannot fully remove. Additional tuning on the same internal split may therefore have less value than clearer ontology definitions and independently collected examples for the affected categories.

The operational taxonomy also separates same-crop disease confusion from cross-crop mistakes. Same-crop condition confusions remain the largest category under both reference and PTE states, which is consistent with fine-grained visual similarity. Fruit-condition involvement grows under the PTE, aligning with the paired decline in fruit recall. These analyses do not prove why an image was misclassified, but they identify where the next data and review effort should be directed.

\subsection{What foundation transfer and teacher guidance do not prove}\label{what-foundation-transfer-and-teacher-guidance-do-not-prove}

DINOv3 supplied the completed reference lineage and the frozen diagnostic representation, but the study does not include a matched non-DINO ConvNeXt control. The evidence therefore shows that a DINOv3-pretrained reference performed strongly, without isolating the improvement attributable to DINOv3. The compact history records hard-label, teacher-logit, and feature-transfer components, yet no matched direct MobileNetV4 baseline survives. The paper accordingly describes a teacher-guided training path without claiming a causal distillation gain. These are natural experiments for an expanded model study, but they are not required to support the current audit-to-runtime result.

\subsection{Internal validity and the next evidence boundary}\label{internal-validity-and-the-next-evidence-boundary}

The reconstructed split addresses known duplicate-family leakage. It does not create a new acquisition domain. Metadata, low-frequency imagery, and borders remain label-predictive, and train, validation, and test originate from the same audited pool. The next decisive evaluation is therefore not another decimal place on the consumed internal test. It is a source-independent smartphone cohort with a prespecified mapping to the 120-class ontology, followed by actual Android measurements of latency, memory, energy, delegate coverage, and thermal behaviour. Until those studies exist, CropCop is best understood as a strongly verified internal recognition system and executable software artifact, not an autonomous field diagnostic.

\section{Limitations}\label{limitations}

\begin{policybox}
\PolicyKicker{Research limitations statement}
\textbf{External validity.} The test is controlled against the audited trusted leakage graph but internal to the reconstructed source pool. Zero observed group crossings do not exclude false negatives in duplicate detection. It cannot establish performance on unseen farms, regions, cultivars, cameras, backgrounds, or acquisition protocols. The shortcut probes make this boundary visible rather than eliminating it.

\vspace{8pt}\textbf{Provenance and ontology.} The surviving archive does not reconstruct complete source URLs, annotation authority, cultivar, geography, severity, or redistribution rights for every image. The 120 operational labels mix disease specificity and visual target types. CropCop is therefore not presented as a legally homogeneous public image release or a definitive botanical taxonomy.

\vspace{8pt}\textbf{Causal model analysis.} The completed evidence lacks a matched direct MobileNetV4 baseline and a matched non-DINO reference control. The study cannot assign the compact result to distillation, feature transfer, or DINOv3 pretraining as isolated causes.

\vspace{8pt}\textbf{Historical mobile configuration.} The selected mobile state, fingerprint, component loss histories, preprocessing contract, and predictions are preserved. The exact optimiser, schedule, distillation temperature, and objective coefficients are not available in a complete human-readable configuration. This limits exact recipe-level reproduction of the historical student run.

\vspace{8pt}\textbf{Training variability.} Each completed final lineage is represented by one archived training seed. Bootstrap intervals quantify uncertainty over the fixed test sample, not variation across retraining. Multi-seed claims are therefore withheld.

\vspace{8pt}\textbf{Runtime scope.} Direct PTE execution proves software-runtime predictions in the archived Kaggle-host environment. The archive does not contain an operator-level XNNPACK delegation/fallback dump, and it does not establish Android latency, memory, energy, delegate fallback, or thermal behaviour. Host-only timings are not used as phone performance evidence.

\vspace{8pt}\textbf{Consumed test.} The final test has been opened for the frozen lineage and the paired analyses reported here. It must not be used to choose new architectures, losses, thresholds, or quantisation recipes. Later studies should rely on validation-only ablations and a new external cohort for consequential comparisons.
\end{policybox}

\section{Data and Artifact Availability}\label{data-and-artifact-availability}

The \href{https://github.com/rana-m-ahmed/ResearchWork-CropCop}{GitHub companion repository} provides the public reproducibility package. It contains the manuscript source, canonical PDF, metric registry, claim-evidence matrix, complete per-class reference and PTE tables, PTQ summaries, paired analyses, deterministic figure-generation code, sanitized notebooks and certificates, model/data cards, and cryptographic identifiers. The repository validator checks manuscript/evidence consistency, relative links, notebook sanitation, both checksum ledgers, citation metadata, and release boundaries. An immutable preprint tag and release manifest will be frozen only after the arXiv identifier is assigned; until then, the repository is a release candidate rather than the archival citation target.

The consolidated image corpus is not distributed with this manuscript because complete image-level provenance and redistribution rights have not been reconstructed. The final PTE and model states are identified by hashes but should not be released until upstream-model and dataset-derived distribution obligations have been reviewed. A release manifest will bind the paper PDF, source archive, repository commit, evidence package, and checksums to the same version.

\section{Intended Use and Safety}\label{intended-use-and-safety}

\begin{policybox}
\PolicyKicker{Intended-use and safety statement}
CropCop is a closed-set research classifier. It always chooses among 120 known labels and has not been evaluated on unsupported crops, novel diseases, non-plant images, or a source-independent field cohort. A high-confidence output can still be wrong, particularly when the image lies outside the frozen distribution. The system must not be presented as autonomous agronomic diagnosis or as a treatment recommender.

\vspace{8pt}Any consequential application should disclose uncertainty, provide abstention and human-review paths, validate the complete image-preprocessing pipeline on target devices, and test performance on independently collected smartphone data. Agronomic decisions should remain subject to qualified expertise and local context.
\end{policybox}

\section{Reproducibility Statement}\label{reproducibility-statement}

\begin{policybox}
\PolicyKicker{Reproducibility statement}
The evidence package preserves the final dataset and class fingerprints, group-safe split counts, reference configuration, checkpoint and artifact hashes, stable row identifiers, raw prediction records, per-class reports, PTQ selection record, and direct PTE outputs. Reference training can be reconstructed from the archived notebook and configuration values reported in Section 4. The paired analyses in Section 6 are derived from the locked rows and are released as machine-readable tables.

\vspace{8pt}The compact model's exact historical optimiser and objective coefficients are the principal reproducibility gap. The manuscript does not hide that gap or fill it by reverse-engineering loss curves. A future reproduced mobile recipe would require a new version identifier and should not be presented as proof of the missing historical settings.
\end{policybox}

\clearpage
\section{AI-Assistance Disclosure}\label{ai-assistance-disclosure}

\begin{policybox}
\PolicyKicker{AI-assistance disclosure}
Generative-AI tools assisted literature organisation, language editing, code inspection, and consistency checking during manuscript preparation. All empirical claims were checked against archived project evidence or explicitly derived from released prediction records. The authors remain responsible for the study design, analyses, claims, and final text.
\end{policybox}

\section{Conclusion}\label{conclusion}

CropCop began as a broad model-building project and became a benchmark-reconstruction project before it could become a model-building project again. Rejecting a partition crossed by 3,233 duplicate relationships produced a frozen 109,107-image, 120-class benchmark with zero crossings among the audited trusted leakage groups. On that benchmark, a DINOv3 ConvNeXt-Tiny reference achieved 98.51\% accuracy and 96.87\% macro-F1, while a compact MobileNetV4 lineage preserved 98.46\% accuracy and 96.27\% macro-F1.

The final result is an identified runtime object rather than an inherited checkpoint number. The 22.60 MiB ExecuTorch/XNNPACK PTE was executed directly, reached 98.46\% accuracy and 96.23\% macro-F1, and changed only six decisions relative to the converted INT8 graph. Paired analysis shows where the remaining cost lies: not in a clear aggregate accuracy loss, but in a modest reduction in class-balanced performance concentrated among low-support and fruit-condition categories. CropCop closes the internal loop from data identity to runtime identity. New farms, phones, and acquisition pipelines remain the next loop to test.

\appendix
\clearpage

\section{Complete per-class results}\label{complete-per-class-results}

The complete 120-class reference and PTE precision, recall, F1, and support tables are distributed as machine-readable files in the companion repository:

\begin{manifestbox}
\begin{itemize}[leftmargin=1.1em,itemsep=4pt,topsep=0pt]
\item \nolinkurl{metrics/reference_per_class_test.csv}
\item \nolinkurl{metrics/pte_runtime_per_class_test.csv}
\item \nolinkurl{evidence/derived/reference_vs_pte_per_class_delta.csv}
\item \nolinkurl{evidence/derived/reference_vs_pte_per_class_counts.csv}
\end{itemize}
\end{manifestbox}

These files preserve class order and full numerical precision. The main paper reports the support-bin summaries and the largest descriptive class-level changes rather than reproducing seven pages of dense tables. Appendix~\ref{operational-class-taxonomy-and-support-inventory} reports the corresponding human-readable crop, label-type, and split-support inventory in the same class order.

\section{Model and dataset identity register}\label{model-and-dataset-identity-register}

\begin{table}[H]
\centering
\caption{Model and dataset identity registry.}
\label{tab:identity-registry}
\begingroup
\sffamily\fontsize{8}{9.8}\selectfont
\setlength{\tabcolsep}{6pt}
\renewcommand{\arraystretch}{1.32}
\begin{tabularx}{\textwidth}{@{}>{\RaggedRight\arraybackslash}p{0.34\textwidth}>{\RaggedRight\arraybackslash}X@{}}
\rowcolor{BenchmarkBlue}
\color{white}\bfseries Object & \color{white}\bfseries Identifier \\
\rowcolor{BenchmarkBlue!8}
Final retained dataset semantic fingerprint & \HashTwo{7c368e6e3d8be3bb3a9a3a5f961075d4}{faa125bcac2e98a3b55e1a1c61f1c523} \\
\rowcolor{SurfaceGrey}
Final dataset build fingerprint & \HashTwo{2d7c237981b8943d9b08a522db048984}{9a4bfe7f488dfb6461489dd36dcdc12b} \\
Dataset manifest SHA-256 & \HashTwo{bdb82211ccc2059153724eea178a1680}{893a6b38ecc243fae484baa91dbf68e2} \\
\rowcolor{SurfaceGrey}
Class-map SHA-256 & \HashTwo{46f7811726c19c42bd7213b2d8178b19}{a5a182a1b763f60a94ee2c0e5f6688d2} \\
Reference checkpoint SHA-256 & \HashTwo{74b4701b8931976c9227845ead50788a}{e47e3596f575f2817b7352a715f53b79} \\
\rowcolor{SurfaceGrey}
Mobile configuration fingerprint & \HashTwo{a75442087ee1c148b1dcd68441f2fd87}{8d8738844532ece4d5606a0038ffb91e} \\
Selected float-state SHA-256 & \HashTwo{b41e76660bcc1042991282aed2ea4067}{ba94677116604b622eda6a8f1eb00c00} \\
\rowcolor{SurfaceGrey}
PTQ selection fingerprint & \HashTwo{bb2ccae4c03f27099cdd1bcbf392676}{355b637df4c9f6c15b49d12884994e20f} \\
Final PTE SHA-256 & \HashTwo{7c70d0f307f0d9578310600913cb8ff1}{71b294ae4de0813f7cd1d6a53628bdf1} \\
\end{tabularx}
\endgroup
\end{table}

\section{Paired state-transition details}\label{paired-state-transition-details}

\subsection{Reference-to-PTE paired bootstrap}\label{reference-to-pte-paired-bootstrap}

\begin{table}[H]
\centering
\caption{Paired bootstrap state-transition summary.}
\label{tab:paired-bootstrap-full}
\begingroup
\sffamily\fontsize{8}{9.8}\selectfont
\setlength{\tabcolsep}{4.5pt}
\renewcommand{\arraystretch}{1.28}
\begin{tabularx}{\textwidth}{@{}>{\RaggedRight\arraybackslash}XRRRRR@{}}
\rowcolor{BenchmarkBlue}
\color{white}\bfseries Metric &
\color{white}\bfseries \shortstack{Reference\\point} &
\color{white}\bfseries \shortstack{PTE\\point} &
\color{white}\bfseries \shortstack{PTE $-$\\reference} &
\color{white}\bfseries \shortstack{95\%\\lower} &
\color{white}\bfseries \shortstack{95\%\\upper} \\
\rowcolor{BenchmarkBlue!8}
Accuracy & 0.985088 & 0.984599 & -0.000489 & -0.002140 & +0.001161 \\
\rowcolor{SurfaceGrey}
Balanced accuracy & 0.966836 & 0.962017 & -0.004819 & -0.010967 & +0.001128 \\
Macro-F1 & 0.968700 & 0.962267 & -0.006433 & -0.012386 & -0.000467 \\
\rowcolor{SurfaceGrey}
Fruit mean recall & 0.859364 & 0.789711 & -0.069653 & -0.124086 & -0.017108 \\
Recall $<100$ & 0.893703 & 0.872686 & -0.021017 & -0.057207 & +0.013036 \\
\rowcolor{SurfaceGrey}
Recall 100--199 & 0.930925 & 0.902687 & -0.028238 & -0.056711 & +0.000623 \\
Recall 200--499 & 0.980353 & 0.983360 & +0.003007 & -0.002884 & +0.008774 \\
\rowcolor{SurfaceGrey}
Recall 500--999 & 0.991013 & 0.990140 & -0.000873 & -0.003038 & +0.001292 \\
Recall $\geq1000$ & 0.985762 & 0.987056 & +0.001294 & -0.001100 & +0.003858 \\
\end{tabularx}
\endgroup
\end{table}

\subsection{Converted-INT8-to-PTE disagreement audit}\label{converted-int8-to-pte-disagreement-audit}

\begin{table}[H]
\centering
\caption{Converted-INT8-to-PTE disagreement audit.}
\label{tab:disagreement-audit-full}
\begingroup
\sffamily\fontsize{7.65}{9.3}\selectfont
\setlength{\tabcolsep}{3.6pt}
\renewcommand{\arraystretch}{1.32}
\begin{tabularx}{\textwidth}{@{}>{\RaggedLeft\arraybackslash}p{0.055\textwidth}>{\RaggedRight\arraybackslash}X>{\RaggedRight\arraybackslash}X>{\RaggedRight\arraybackslash}X>{\RaggedLeft\arraybackslash}p{0.09\textwidth}>{\RaggedLeft\arraybackslash}p{0.085\textwidth}@{}}
\rowcolor{BenchmarkBlue}
\color{white}\bfseries Row ID & \color{white}\bfseries True class & \color{white}\bfseries Converted INT8 & \color{white}\bfseries PTE & \color{white}\bfseries \shortstack{INT8\\confidence} & \color{white}\bfseries \shortstack{PTE\\confidence} \\
\rowcolor{BenchmarkBlue!8}
1455 & \nolinkurl{banana_healthy_fruit} & \nolinkurl{banana_healthy_fruit} $\checkmark$ & \nolinkurl{banana_disease_fruit} $\times$ & 0.479 & 0.477 \\
\rowcolor{SurfaceGrey}
3891 & \nolinkurl{corn_northern_leaf_blight} & \nolinkurl{corn_northern_leaf_blight} $\checkmark$ & \nolinkurl{wheat_leaf_rust} $\times$ & 0.303 & 0.358 \\
6292 & \nolinkurl{nutrient_deficiency} & \nolinkurl{pear_black_spot} $\times$ & \nolinkurl{nutrient_deficiency} $\checkmark$ & 0.486 & 0.468 \\
\rowcolor{SurfaceGrey}
11743 & \nolinkurl{sugarcane_yellow_leaf} & \nolinkurl{sugarcane_mosaic} $\times$ & \nolinkurl{sugarcane_yellow_leaf} $\checkmark$ & 0.530 & 0.472 \\
12361 & \nolinkurl{tomato_early_blight} & \nolinkurl{tomato_septoria_leaf_spot} $\times$ & \nolinkurl{tomato_early_blight} $\checkmark$ & 0.481 & 0.477 \\
\rowcolor{SurfaceGrey}
12750 & \nolinkurl{tomato_healthy} & \nolinkurl{apple_healthy} $\times$ & \nolinkurl{cherry_healthy} $\times$ & 0.262 & 0.267 \\
\end{tabularx}
\endgroup
\end{table}

\subsection{Reference-to-PTE operational error taxonomy}\label{reference-to-pte-operational-error-taxonomy}

\begin{table}[H]
\centering
\caption{Reference-to-PTE operational error taxonomy.}
\label{tab:error-taxonomy-full}
\begingroup
\TableText
\setlength{\tabcolsep}{7pt}
\renewcommand{\arraystretch}{1.3}
\begin{tabularx}{\textwidth}{@{}>{\RaggedRight\arraybackslash}XRRR@{}}
\rowcolor{BenchmarkBlue}
\color{white}\bfseries Operational error category & \color{white}\bfseries Reference errors & \color{white}\bfseries PTE errors & \color{white}\bfseries Change \\
\rowcolor{BenchmarkBlue!8}
Cross-crop condition confusion & 52 & 48 & -4 \\
\rowcolor{SurfaceGrey}
Cross-crop healthy/condition confusion & 6 & 9 & +3 \\
Fruit-condition label involved & 47 & 64 & +17 \\
\rowcolor{SurfaceGrey}
Generic \Code{plant\_healthy} involved & 22 & 12 & -10 \\
Same-crop condition confusion & 88 & 90 & +2 \\
\rowcolor{SurfaceGrey}
Same-crop healthy/condition confusion & 29 & 29 & +0 \\
\end{tabularx}
\endgroup
\end{table}

\section{Runtime and quantisation environment}\label{runtime-and-quantisation-environment}

\begin{table}[H]
\centering
\caption{Archived runtime and quantisation environment.}
\label{tab:runtime-environment-full}
\begingroup
\TableText
\setlength{\tabcolsep}{6pt}
\renewcommand{\arraystretch}{1.28}
\begin{tabularx}{\textwidth}{@{}>{\RaggedRight\arraybackslash}p{0.36\textwidth}>{\RaggedRight\arraybackslash}X@{}}
\rowcolor{BenchmarkBlue}
\color{white}\bfseries Field & \color{white}\bfseries Archived value \\
\rowcolor{BenchmarkBlue!8}
Evaluation context & Kaggle-host CPU software-runtime evaluation; not Android device evidence \\
\rowcolor{SurfaceGrey}
Python & \Code{3.12.13} \\
PyTorch / torchvision & \Code{2.12.1 / 0.27.1} \\
\rowcolor{SurfaceGrey}
ExecuTorch / torchao & \Code{1.3.1 / 0.17.0} \\
timm / FlatBuffers & \Code{1.0.26 / 25.12.19} \\
\rowcolor{SurfaceGrey}
Backend / input & XNNPACK / \Code{[1, 3, 256, 256]} \\
Requested quantisation scope & \Code{full} \\
\rowcolor{SurfaceGrey}
Activation / weight scheme & dynamic activation / per-channel weights \\
Explicit module exclusions & \Code{[]} \\
\rowcolor{PassGreen!7}
Coverage gate & \StatusPass \\
Lowered backend modules / delegate tokens & \Code{1 / 1} \\
\rowcolor{WarningAmber!7}
Unarchived host details & CPU model, thread count, and OS build \\
\rowcolor{WarningAmber!7}
Physical Android evaluation & Not evaluated \\
\end{tabularx}
\endgroup
\end{table}

\section{Operational class taxonomy and support inventory}\label{operational-class-taxonomy-and-support-inventory}

Table~\ref{tab:class-taxonomy-inventory} preserves the authoritative class order and reports the frozen train and test support for every operational label. Class labels follow \Code{data\_card/class\_to\_idx.json}; supports follow \Code{data\_card/class\_distribution.csv} and agree row-for-row with both per-class metric tables and the two reference-to-PTE derived tables.

\begingroup
\sffamily\fontsize{7.45}{9.2}\selectfont
\setlength{\tabcolsep}{3.7pt}
\renewcommand{\arraystretch}{1.22}
\begin{longtable}{@{}>{\RaggedRight\arraybackslash}p{0.31\linewidth}>{\RaggedRight\arraybackslash}p{0.13\linewidth}>{\RaggedRight\arraybackslash}p{0.285\linewidth}>{\RaggedLeft\arraybackslash}p{0.085\linewidth}>{\RaggedLeft\arraybackslash}p{0.075\linewidth}@{}}
\caption{Operational class taxonomy and split support.}\label{tab:class-taxonomy-inventory}\\
\rowcolor{BenchmarkBlue}
\color{white}\bfseries Operational class &
\color{white}\bfseries Crop/species &
\color{white}\bfseries Label type &
\color{white}\bfseries Train &
\color{white}\bfseries Test \\
\endfirsthead
\multicolumn{5}{r}{\sffamily\fontsize{7.5}{9.3}\selectfont\color{MutedSlate}Table~\thetable\ --- continued}\\[4pt]
\rowcolor{BenchmarkBlue}
\color{white}\bfseries Operational class &
\color{white}\bfseries Crop/species &
\color{white}\bfseries Label type &
\color{white}\bfseries Train &
\color{white}\bfseries Test \\
\endhead
\bottomrule
\endfoot
\bottomrule
\endlastfoot
\rowcolor{BenchmarkBlue!8}
\nolinkurl{aloevera_healthy} & \nolinkurl{aloevera} & Generic healthy state & 63 & 14 \\
\rowcolor{SurfaceGrey}
\nolinkurl{amla_healthy} & \nolinkurl{amla} & Generic healthy state & 49 & 10 \\
\nolinkurl{apple_black_rot} & \nolinkurl{apple} & Specific named disease & 1,426 & 306 \\
\rowcolor{SurfaceGrey}
\nolinkurl{apple_black_spot} & \nolinkurl{apple} & Specific named disease & 218 & 47 \\
\nolinkurl{apple_brown_spot} & \nolinkurl{apple} & Specific named disease & 696 & 149 \\
\rowcolor{SurfaceGrey}
\nolinkurl{apple_cedar_rust} & \nolinkurl{apple} & Specific named disease & 202 & 43 \\
\nolinkurl{apple_disease_fruit} & \nolinkurl{apple} & Broad condition category & 197 & 42 \\
\rowcolor{SurfaceGrey}
\nolinkurl{apple_healthy} & \nolinkurl{apple} & Generic healthy state & 2,143 & 459 \\
\nolinkurl{apple_scab} & \nolinkurl{apple} & Specific named disease & 399 & 86 \\
\rowcolor{SurfaceGrey}
\nolinkurl{apricot_blight} & \nolinkurl{apricot} & Specific named disease & 165 & 35 \\
\nolinkurl{apricot_healthy} & \nolinkurl{apricot} & Generic healthy state & 418 & 89 \\
\rowcolor{SurfaceGrey}
\nolinkurl{apricot_shot_hole} & \nolinkurl{apricot} & Specific named disease & 546 & 117 \\
\nolinkurl{banana_cordana} & \nolinkurl{banana} & Specific named disease & 185 & 39 \\
\rowcolor{SurfaceGrey}
\nolinkurl{banana_disease_fruit} & \nolinkurl{banana} & Broad condition category & 51 & 11 \\
\nolinkurl{banana_healthy_fruit} & \nolinkurl{banana} & Generic healthy state & 89 & 19 \\
\rowcolor{SurfaceGrey}
\nolinkurl{banana_healthy_leaf} & \nolinkurl{banana} & Generic healthy state & 68 & 15 \\
\nolinkurl{bean_fungal_disease} & \nolinkurl{bean} & Broad condition category & 495 & 106 \\
\rowcolor{SurfaceGrey}
\nolinkurl{bean_healthy} & \nolinkurl{bean} & Generic healthy state & 560 & 120 \\
\nolinkurl{bean_rust} & \nolinkurl{bean} & Specific named disease & 324 & 69 \\
\rowcolor{SurfaceGrey}
\nolinkurl{bean_shot_hole} & \nolinkurl{bean} & Specific named disease & 250 & 54 \\
\nolinkurl{blueberry_healthy} & \nolinkurl{blueberry} & Generic healthy state & 894 & 191 \\
\rowcolor{SurfaceGrey}
\nolinkurl{cherry_brown_spot} & \nolinkurl{cherry} & Specific named disease & 642 & 138 \\
\nolinkurl{cherry_healthy} & \nolinkurl{cherry} & Generic healthy state & 772 & 165 \\
\rowcolor{SurfaceGrey}
\nolinkurl{cherry_leaf_scorch} & \nolinkurl{cherry} & Specific named disease & 632 & 135 \\
\nolinkurl{cherry_powdery_mildew} & \nolinkurl{cherry} & Specific named disease & 576 & 124 \\
\rowcolor{SurfaceGrey}
\nolinkurl{cherry_purple_leaf_spot} & \nolinkurl{cherry} & Specific named disease & 665 & 143 \\
\nolinkurl{cherry_shot_hole} & \nolinkurl{cherry} & Specific named disease & 275 & 59 \\
\rowcolor{SurfaceGrey}
\nolinkurl{citrus_canker_leaf} & \nolinkurl{citrus} & Specific named disease & 117 & 25 \\
\nolinkurl{citrus_fruit_disease} & \nolinkurl{citrus} & Broad condition category & 87 & 19 \\
\rowcolor{SurfaceGrey}
\nolinkurl{citrus_healthy_fruit} & \nolinkurl{citrus} & Generic healthy state & 128 & 28 \\
\nolinkurl{citrus_healthy_leaf} & \nolinkurl{citrus} & Generic healthy state & 96 & 21 \\
\rowcolor{SurfaceGrey}
\nolinkurl{corn_common_rust} & \nolinkurl{corn} & Specific named disease & 1,556 & 334 \\
\nolinkurl{corn_fungal_leaf} & \nolinkurl{corn} & Broad condition category & 147 & 32 \\
\rowcolor{SurfaceGrey}
\nolinkurl{corn_gray_leaf_spot} & \nolinkurl{corn} & Specific named disease & 674 & 144 \\
\nolinkurl{corn_healthy} & \nolinkurl{corn} & Generic healthy state & 1,555 & 333 \\
\rowcolor{SurfaceGrey}
\nolinkurl{corn_holcus_leaf_spot} & \nolinkurl{corn} & Specific named disease & 219 & 47 \\
\nolinkurl{corn_northern_leaf_blight} & \nolinkurl{corn} & Specific named disease & 1,577 & 338 \\
\rowcolor{SurfaceGrey}
\nolinkurl{crape_jasmine_healthy} & \nolinkurl{crape_jasmine} & Generic healthy state & 32 & 7 \\
\nolinkurl{fig_blight} & \nolinkurl{fig} & Specific named disease & 514 & 110 \\
\rowcolor{SurfaceGrey}
\nolinkurl{fig_brown_spot} & \nolinkurl{fig} & Specific named disease & 261 & 57 \\
\nolinkurl{fig_healthy} & \nolinkurl{fig} & Generic healthy state & 400 & 86 \\
\rowcolor{SurfaceGrey}
\nolinkurl{fig_rust} & \nolinkurl{fig} & Specific named disease & 331 & 71 \\
\nolinkurl{grape_anthracnose} & \nolinkurl{grape} & Specific named disease & 675 & 144 \\
\rowcolor{SurfaceGrey}
\nolinkurl{grape_black_rot} & \nolinkurl{grape} & Specific named disease & 694 & 149 \\
\nolinkurl{grape_brown_spot} & \nolinkurl{grape} & Specific named disease & 432 & 93 \\
\rowcolor{SurfaceGrey}
\nolinkurl{grape_downy_mildew} & \nolinkurl{grape} & Specific named disease & 418 & 89 \\
\nolinkurl{grape_esca} & \nolinkurl{grape} & Specific named disease & 775 & 166 \\
\rowcolor{SurfaceGrey}
\nolinkurl{grape_healthy} & \nolinkurl{grape} & Generic healthy state & 1,239 & 266 \\
\nolinkurl{grape_leaf_blight} & \nolinkurl{grape} & Specific named disease & 603 & 129 \\
\rowcolor{SurfaceGrey}
\nolinkurl{grape_mites} & \nolinkurl{grape} & Specific named disease\textsuperscript{\textdagger} & 282 & 60 \\
\nolinkurl{grape_powdery_mildew} & \nolinkurl{grape} & Specific named disease & 803 & 172 \\
\rowcolor{SurfaceGrey}
\nolinkurl{grape_shot_hole} & \nolinkurl{grape} & Specific named disease & 557 & 120 \\
\nolinkurl{guava_disease_fruit} & \nolinkurl{guava} & Broad condition category & 110 & 23 \\
\rowcolor{SurfaceGrey}
\nolinkurl{guava_healthy_fruit} & \nolinkurl{guava} & Generic healthy state & 35 & 7 \\
\nolinkurl{guava_healthy_leaf} & \nolinkurl{guava} & Generic healthy state & 105 & 22 \\
\rowcolor{SurfaceGrey}
\nolinkurl{guava_red_rust} & \nolinkurl{guava} & Specific named disease & 63 & 13 \\
\nolinkurl{hibiscus_healthy} & \nolinkurl{hibiscus} & Generic healthy state & 58 & 13 \\
\rowcolor{SurfaceGrey}
\nolinkurl{loquat_healthy} & \nolinkurl{loquat} & Generic healthy state & 334 & 71 \\
\nolinkurl{loquat_leaf_spot} & \nolinkurl{loquat} & Specific named disease & 437 & 93 \\
\rowcolor{SurfaceGrey}
\nolinkurl{mango_bacterial_canker} & \nolinkurl{mango} & Specific named disease & 273 & 59 \\
\nolinkurl{mango_disease_fruit} & \nolinkurl{mango} & Broad condition category & 143 & 30 \\
\rowcolor{SurfaceGrey}
\nolinkurl{mango_healthy} & \nolinkurl{mango} & Generic healthy state & 58 & 12 \\
\nolinkurl{mango_healthy_fruit} & \nolinkurl{mango} & Generic healthy state & 227 & 49 \\
\rowcolor{SurfaceGrey}
\nolinkurl{mango_healthy_leaf} & \nolinkurl{mango} & Generic healthy state & 212 & 46 \\
\nolinkurl{money_plant_healthy} & \nolinkurl{money_plant} & Generic healthy state & 63 & 14 \\
\rowcolor{SurfaceGrey}
\nolinkurl{nutrient_deficiency} & \nolinkurl{unspecified} & Broad condition category & 83 & 18 \\
\nolinkurl{orange_citrus_greening} & \nolinkurl{orange} & Specific named disease & 2,885 & 618 \\
\rowcolor{SurfaceGrey}
\nolinkurl{papaya_fruit_disease} & \nolinkurl{papaya} & Broad condition category & 73 & 16 \\
\nolinkurl{papaya_fruit_healthy} & \nolinkurl{papaya} & Generic healthy state & 96 & 20 \\
\rowcolor{SurfaceGrey}
\nolinkurl{papaya_healthy_leaf} & \nolinkurl{papaya} & Generic healthy state & 208 & 45 \\
\nolinkurl{papaya_ring_spot} & \nolinkurl{papaya} & Specific named disease & 471 & 101 \\
\rowcolor{SurfaceGrey}
\nolinkurl{peach_bacterial_spot} & \nolinkurl{peach} & Specific named disease & 1,285 & 276 \\
\nolinkurl{peach_healthy} & \nolinkurl{peach} & Generic healthy state & 264 & 57 \\
\rowcolor{SurfaceGrey}
\nolinkurl{pear_black_spot} & \nolinkurl{pear} & Specific named disease & 553 & 119 \\
\nolinkurl{pear_fire_blight} & \nolinkurl{pear} & Specific named disease & 250 & 53 \\
\rowcolor{SurfaceGrey}
\nolinkurl{pear_healthy} & \nolinkurl{pear} & Generic healthy state & 478 & 103 \\
\nolinkurl{pepper_bacterial_spot} & \nolinkurl{pepper} & Specific named disease & 586 & 125 \\
\rowcolor{SurfaceGrey}
\nolinkurl{pepper_healthy} & \nolinkurl{pepper} & Generic healthy state & 846 & 181 \\
\nolinkurl{persimmon_brown_spot} & \nolinkurl{persimmon} & Specific named disease & 432 & 92 \\
\rowcolor{SurfaceGrey}
\nolinkurl{plant_healthy} & \nolinkurl{unspecified} & Generic healthy state & 229 & 46 \\
\nolinkurl{potato_early_blight} & \nolinkurl{potato} & Specific named disease & 766 & 164 \\
\rowcolor{SurfaceGrey}
\nolinkurl{potato_healthy} & \nolinkurl{potato} & Generic healthy state & 105 & 23 \\
\nolinkurl{potato_late_blight} & \nolinkurl{potato} & Specific named disease & 763 & 163 \\
\rowcolor{SurfaceGrey}
\nolinkurl{raspberry_healthy} & \nolinkurl{raspberry} & Generic healthy state & 264 & 57 \\
\nolinkurl{rice_blast} & \nolinkurl{rice} & Specific named disease & 409 & 88 \\
\rowcolor{SurfaceGrey}
\nolinkurl{rice_brown_spot} & \nolinkurl{rice} & Specific named disease & 425 & 91 \\
\nolinkurl{rice_healthy} & \nolinkurl{rice} & Generic healthy state & 420 & 90 \\
\rowcolor{SurfaceGrey}
\nolinkurl{rose_healthy} & \nolinkurl{rose} & Generic healthy state & 20 & 4 \\
\nolinkurl{soybean_healthy} & \nolinkurl{soybean} & Generic healthy state & 2,750 & 589 \\
\rowcolor{SurfaceGrey}
\nolinkurl{squash_powdery_mildew} & \nolinkurl{squash} & Specific named disease & 1,108 & 237 \\
\nolinkurl{strawberry_healthy} & \nolinkurl{strawberry} & Generic healthy state & 306 & 65 \\
\rowcolor{SurfaceGrey}
\nolinkurl{strawberry_leaf_scorch} & \nolinkurl{strawberry} & Specific named disease & 618 & 132 \\
\nolinkurl{sugarcane_bacterial_blight} & \nolinkurl{sugarcane} & Specific named disease & 2,974 & 637 \\
\rowcolor{SurfaceGrey}
\nolinkurl{sugarcane_healthy} & \nolinkurl{sugarcane} & Generic healthy state & 1,191 & 256 \\
\nolinkurl{sugarcane_mosaic} & \nolinkurl{sugarcane} & Specific named disease & 874 & 188 \\
\rowcolor{SurfaceGrey}
\nolinkurl{sugarcane_red_rot} & \nolinkurl{sugarcane} & Specific named disease & 1,636 & 350 \\
\nolinkurl{sugarcane_rust} & \nolinkurl{sugarcane} & Specific named disease & 1,466 & 314 \\
\rowcolor{SurfaceGrey}
\nolinkurl{sugarcane_yellow_leaf} & \nolinkurl{sugarcane} & Specific named disease & 1,203 & 258 \\
\nolinkurl{tomato_bacterial_spot} & \nolinkurl{tomato} & Specific named disease & 2,126 & 455 \\
\rowcolor{SurfaceGrey}
\nolinkurl{tomato_early_blight} & \nolinkurl{tomato} & Specific named disease & 1,403 & 300 \\
\nolinkurl{tomato_fusarium_wilt} & \nolinkurl{tomato} & Specific named disease & 140 & 30 \\
\rowcolor{SurfaceGrey}
\nolinkurl{tomato_healthy} & \nolinkurl{tomato} & Generic healthy state & 1,941 & 416 \\
\nolinkurl{tomato_late_blight} & \nolinkurl{tomato} & Specific named disease & 1,854 & 397 \\
\rowcolor{SurfaceGrey}
\nolinkurl{tomato_leaf_curl} & \nolinkurl{tomato} & Specific named disease & 553 & 119 \\
\nolinkurl{tomato_leaf_miner} & \nolinkurl{tomato} & Specific named disease\textsuperscript{\textdagger} & 606 & 130 \\
\rowcolor{SurfaceGrey}
\nolinkurl{tomato_leaf_mold} & \nolinkurl{tomato} & Specific named disease & 1,351 & 290 \\
\nolinkurl{tomato_mosaic_virus} & \nolinkurl{tomato} & Specific named disease & 240 & 51 \\
\rowcolor{SurfaceGrey}
\nolinkurl{tomato_septoria_leaf_spot} & \nolinkurl{tomato} & Specific named disease & 1,710 & 367 \\
\nolinkurl{tomato_spider_mites} & \nolinkurl{tomato} & Specific named disease\textsuperscript{\textdagger} & 1,281 & 275 \\
\rowcolor{SurfaceGrey}
\nolinkurl{tomato_target_spot} & \nolinkurl{tomato} & Specific named disease & 982 & 211 \\
\nolinkurl{tomato_verticillium_wilt} & \nolinkurl{tomato} & Specific named disease & 170 & 37 \\
\rowcolor{SurfaceGrey}
\nolinkurl{tomato_yellow_leaf_curl} & \nolinkurl{tomato} & Specific named disease & 2,840 & 608 \\
\nolinkurl{walnut_anthracnose} & \nolinkurl{walnut} & Specific named disease & 436 & 93 \\
\rowcolor{SurfaceGrey}
\nolinkurl{walnut_blotch} & \nolinkurl{walnut} & Specific named disease & 852 & 183 \\
\nolinkurl{walnut_gall_mite} & \nolinkurl{walnut} & Specific named disease\textsuperscript{\textdagger} & 269 & 58 \\
\rowcolor{SurfaceGrey}
\nolinkurl{walnut_healthy} & \nolinkurl{walnut} & Generic healthy state & 413 & 88 \\
\nolinkurl{walnut_shot_hole} & \nolinkurl{walnut} & Specific named disease & 623 & 133 \\
\rowcolor{SurfaceGrey}
\nolinkurl{wheat_healthy} & \nolinkurl{wheat} & Generic healthy state & 623 & 133 \\
\nolinkurl{wheat_leaf_rust} & \nolinkurl{wheat} & Specific named disease & 395 & 85 \\
\rowcolor{SurfaceGrey}
\nolinkurl{wheat_stripe_rust} & \nolinkurl{wheat} & Specific named disease & 241 & 51 \\\end{longtable}
\endgroup

\ObjectNote{Crop/species is inferred from the literal label prefix; \Code{nutrient\_deficiency} and \Code{plant\_healthy} are crop-unspecified. Label types follow the paper's three-way naming logic: labels containing \Code{healthy} are generic healthy states; generic fruit/leaf disease, fungal disease/leaf, and nutrient-deficiency labels are broad condition categories; the remaining non-healthy labels are assigned to the specific named disease category. \textsuperscript{\textdagger}\Code{grape\_mites}, \Code{tomato\_leaf\_miner}, \Code{tomato\_spider\_mites}, and \Code{walnut\_gall\_mite} are named pest or arthropod conditions and are flagged because the requested three-way scheme has no separate pest category. \Code{rice\_neck\_blast} is not one of the 120 operational classes; it was a one-image candidate category quarantined before the final freeze.}

\FloatBarrier
{\small\bibliographystyle{unsrtnat}\bibliography{references}}
\end{document}